%% file: main.tex
\documentclass[10pt,twocolumn,letterpaper]{article}

\usepackage[cvprfinal,pagenumbers]{cvpr} % To produce the CAMERA-READY version
\usepackage{graphicx}
\usepackage{amsmath}
\usepackage{amssymb}
\usepackage{booktabs}
\usepackage{mathtools}
\usepackage{iac_pkg}
\usepackage{lipsum}
\usepackage{amssymb}% http://ctan.org/pkg/amssymb
\usepackage{pifont}% http://ctan.org/pkg/pifont
\usepackage{multirow}
\usepackage{tabularx, booktabs}
\usepackage[htt]{hyphenat}
\usepackage{color}
\usepackage{xspace}
\usepackage{cite}
\usepackage{overpic}
\usepackage{arydshln}
\usepackage{subcaption}
\usepackage{xcolor}

\definecolor{citecolor}{RGB}{34,139,34}
\definecolor{lightred}{RGB}{255,100,100}
\definecolor{cell_bisque}{rgb}{1.0, 0.89, 0.77}
\definecolor{cell_blond}{rgb}{0.98, 0.94, 0.75}
\definecolor{cell_blue}{RGB}{155, 187, 228}
\definecolor{princetonorange}{rgb}{1.0, 0.56, 0.0}
\definecolor{pinkpearl}{rgb}{0.91, 0.67, 0.81}
\definecolor{mossgreen}{rgb}{0.68, 0.87, 0.68}

\newcommand{\Paragraph}[1]{\vspace{-0mm}\noindent\textbf{#1.}\hspace{0mm}}
\newcommand{\Section}[1]{\vspace{-1mm} \section{#1} \vspace{-0mm}}
\newcommand{\SubSection}[1]{\vspace{-1mm} \subsection{#1} \vspace{-0mm}}
\newcommand{\SubSubSection}[1]{\vspace{-1mm} \subsubsection{#1} \vspace{-0mm}}

\usepackage{hyperref}
\hypersetup{pagebackref=true,breaklinks=true,letterpaper=true,colorlinks,bookmarks=false,citecolor=cyan}

\usepackage[capitalize]{cleveref}

\def\cvprPaperID{10047} % *** Enter the CVPR Paper ID here
\def\confName{CVPR}
\def\confYear{2023}
\begin{document}

\title{Hierarchical Fine-Grained Image Forgery Detection and Localization}

\author{Xiao Guo$^{1}$, Xiaohong Liu$^{2}$, Zhiyuan Ren$^{1}$, Steven Grosz$^{1}$,  Iacopo Masi$^{3}$, Xiaoming Liu$^{1}$\\
$^{1}$ Michigan State University $^{2}$ Shanghai Jiao Tong University $^{3}$ Sapienza University of Rome \\
{\tt\small \{guoxia11, renzhiy1, groszst\}@msu.edu, xiaohongliu@sjtu.edu.cn,} \\{\tt\small masi@di.uniroma1.it, liuxm@msu.edu}\\
% {\tt\small masi@di.uniroma1.it  \{renzhiy1, groszst, liuxm\}@msu.edu}
}

\maketitle

\input{section/00_abstract.tex}
\input{section/01_introduction.tex}
\input{section/02_related_work.tex}
\input{section/03_00_model.tex}
\input{section/03_01_benchmark.tex}

\input{section/04_00_experiment.tex}
\input{section/04_01_experiment_ablation}
\input{section/05_conclusion.tex}

% %------------------------------------------------------------------------
% \clearpage
\section{Supplementary}
In this supplementary material, we include many details of our work: 1) the details of the proposed HiFi-IFDL dataset; 2) The generalization performance against images generated from unseen forgery methods and real images in the unseen domain; 3) the HiFi-Net performance against different types of post-processing in the image editing domain; 4) the complete HiFi-Net performance on the DFFD dataset~\cite{stehouwer2019detection}; 5) we offer the forgery attribute classification results on seen and unseen forgery attributes; 6). the detailed implementation of the proposed HiFi-Net.
\input{section/10_appendix.tex}
{\small
\bibliographystyle{ieee_fullname}
\bibliography{00_dfd}
}
\end{document}

%% file: section/00_abstract.tex
%%%%%%%%% ABSTRACT
\begin{abstract}
% Unlike the previous image forgery detection and localization (IFDL) work which specializes in either image editing or CNN-synthesized domains, we present a unified method for IFDL, through a hierarchical fine-grained formulation. We first 
% It is challenging to develop an effective unified algorithm for image forgery detection and localization (IFDL), in both CNN-synthesized and image editing domains, since the forgery attributes differ largely between different forgery methods. 
Differences in forgery attributes of images generated in CNN-synthesized and image-editing domains are large, and such differences make a unified image forgery detection and localization (IFDL) challenging.
To this end, we present a hierarchical fine-grained formulation for IFDL representation learning. Specifically, we first represent forgery attributes of a manipulated image with multiple labels at different levels. Then we perform fine-grained classification at these levels using the hierarchical dependency between them. As a result, the algorithm is encouraged to learn both comprehensive features and inherent hierarchical nature of different forgery attributes, thereby improving the IFDL representation. 
Our proposed IFDL framework contains three components: multi-branch feature extractor, localization and classification modules. 
Each branch of the feature extractor learns to classify forgery attributes at one level, while localization and classification modules segment the pixel-level forgery region and detect image-level forgery, respectively. Lastly, we construct a hierarchical fine-grained dataset to facilitate our study. We demonstrate the effectiveness of our method on $7$ different benchmarks, for both tasks of IFDL and forgery attribute classification. Our source code and dataset can be found: \href{https://github.com/CHELSEA234/HiFi_IFDL}{github.com/CHELSEA234/HiFi-IFDL}.
\end{abstract}
\vspace{-5mm}

%% file: section/01_introduction.tex
\Section{Introduction}\label{sec:intro}
\input{section/figure_table_latex/figure_overview_0} 
\input{section/figure_table_latex/figure_overview} 
Chaotic and pervasive multimedia information sharing offers better means for spreading misinformation~\cite{cnn_internet}, and the forged image content could, in principle, sustain recent ``infodemics''~\cite{infodemics}. Firstly, CNN-synthesized images made extraordinary leaps culminating in recent synthesis methods---Dall$\cdot$E~\cite{ramesh2022hierarchical} or Google ImageN~\cite{saharia2022photorealistic}---based on diffusion models (DDPM)~\cite{ho2020denoising_ddpm}, which even generates realistic videos from text~\cite{singer2022make,ho2022imagen}. 
Secondly, the availability of image editing toolkits produced a substantially low-cost access to image forgery or tampering (\textit{e.g.}, splicing and inpainting). In response to such an issue of image forgery, the computer vision community has made considerable efforts, which however branch separately into two directions: detecting either CNN synthesis~\cite{zhangxue2019detecting,wang2020cnn,stehouwer2019detection}, or conventional image editing~\cite{wu2019mantra,hu2020span,liu2022pscc,dong2022mvss,wang2022objectformer}. As a result, these methods may be ineffective when deploying to real-life scenarios, where forged images can possibly be generated from either CNN-sythensized or image-editing domains.

To push the frontier of image forensics~\cite{sencar2022multimedia}, we study the image forgery detection and localization problem (IFDL)---Fig.~\ref{fig_mani_overview}\textcolor{red}{a}---regardless of the forgery method domains, \textit{i.e.}, CNN-synthesized or image editing. 
It is challenging to develop a unified algorithm for two domains, as images, generated by different forgery methods, differ largely from each other in terms of various forgery attributes. For example, a forgery attribute can indicate whether a forged image is fully synthesized or partially manipulated, or whether the forgery method used is the diffusion model generating images from the Gaussian noise, or an image editing process that splices two images via Poisson editing~\cite{perez2003poisson}. 
Therefore, to model such complex forgery attributes, we first represent forgery attribute of each forged image with multiple labels at different levels. Then, we present a hierarchical fine-grained formulation for IFDL, which requires the algorithm to classify fine-grained forgery attributes of each image at different levels, via the inherent hierarchical nature of different forgery attributes.

Fig.~\ref{fig_overview_2}\textcolor{red}{a} shows the interpretation of the forgery attribute with a hierarchy, which evolves from the general forgery attribute, fully-synthesized vs partial-manipulated, to specific individual forgery methods, such as DDPM~\cite{ho2020denoising_ddpm} and DDIM~\cite{song2020denoising_ddim}.
% , in which we not only include the aforementioned forgery attributes, also we try to explore whether the forgery or manipulation is conditioned on certain source (\textit{e.g.}, category labels or implicit styles of the source image), and classify each individual forgery method, such as DDPM~\cite{ho2020denoising_ddpm} and DDIM~\cite{song2020denoising_ddim}. 
Then, given an input image, our method performs fine-grained forgery attribute classification at different levels (see Fig.~\ref{fig_overview_2}\textcolor{red}{b}). 
The image-level forgery detection benefits from this hierarchy as the fine-grained classification learns the comprehensive IFDL representation to differentiate individual forgery methods. Also, for the pixel-level localization, the fine-grained classification features can serve as a prior to improve the localization. 
This holds since the distribution of the forgery area is prominently correlated with  forgery methods, as depicted in Fig.~\ref{fig_mani_overview}\textcolor{red}{b}.
%%% Iac: i see what you are saying.
%% you want to condition the localization over the predicted category
%% that is, if the network predicts
%% stayleGAN, fake patterns are in entire image; if deepfakes look into face area etc. cool, i like it.

In Fig.~\ref{fig_overview_2}\textcolor{red}{c}, we leverage the hierarchical dependency between forgery attributes in fine-grained classification.
Each node's classification probability is conditioned on the path from the root to itself. 
For example, the classification probability at a node of \texttt{DDPM} is conditioned on the classification probability of all nodes in the path of \texttt{Forgery$\rightarrow$ Fully Synthesis$\rightarrow$Diffusion$\rightarrow$Unconditional$\rightarrow$DDPM}. This differs to prior work~\cite{wu2019mantra,marra2018detection,yu2019attributing_image_attribute,marra2019gans} which assume a ``flat'' structure in which attributes are mutually exclusive. Predicting the entire hierarchical path helps understanding forgery attributes from the coarse to fine, thereby capturing dependencies among individual forgery attributes.%% IAC comment: if we need space this last paragrah can be removed. It is the same sentence we said before.

To this end, we propose Hierarchical Fine-grained Network (HiFi-Net).
HiFi-Net has three components: multi-branch feature extractor, localization module and detection module. 
Each branch of the multi-branch extractor classifies images at one forgery attribute level. 
The localization module generates the forgery mask with the help of a deep-metric learning based objective, which improves the separation between real and forged pixels. 
The classification module first overlays the forgery mask with the  input image and obtain a masked image where only forged pixels remain. Then, we use partial convolution to process masked images, which further helps learn IFDL representations.

Lastly, to faciliate our study of the hierarchical fine-grained formulation, we construct a new dataset, termed Hierarchical Fine-grained (HiFi) IFDL dataset. It contains $13$ forgery methods, which are either latest CNN-synthesized methods or representative image editing methods. HiFi-IFDL dataset also induces a hierarchical structure on forgery categories to enable learning a classifier for various forgery attributes.
% such as classic image editing (\textit{e.g.}, splicing and inpainting), diffusion-based full synthesis, and CNN-based partial manipulations. 
Each forged image is also paired with a high-resolution ground truth forgery mask for the localization task.
% Also, TaxIMDL dataset provides authentic images from multiple domains like scenes, objects, and faces, unlike recent work that focuses only on facial images~\cite{stehouwer2019detection,he2021forgerynet}.
In summary, our contributions are as follows:

$\diamond$ We study the task of image forgery detection and localization (IFDL) for both image editing and CNN-synthesized domains. We propose a hierarchical fine-grained formulation to learn a comprehensive representation for IFDL and forgery attribute classification.

$\diamond$ We propose a IFDL algorithm, named HiFi-Net, which not only performs well on forgery detection and localization, also identifies a diverse spectrum of forgery attributes.
% Such architecture that is trained for IMDL can predict the manipulation attribution.

$\diamond$ We construct a new dataset (HiFi-IFDL) to facilitate the hierarchical fine-grained IFDL study. When evaluating  on $7$ benchmarks, our method outperforms the state of the art (SoTA) on the tasks of IFDL, and achieve a competitive performance on the forgery attribute classifications.

%% file: section/figure_table_latex/figure_overview_0.tex
\begin{figure}[t]
    \centering
    \begin{overpic}[width=0.5\textwidth]{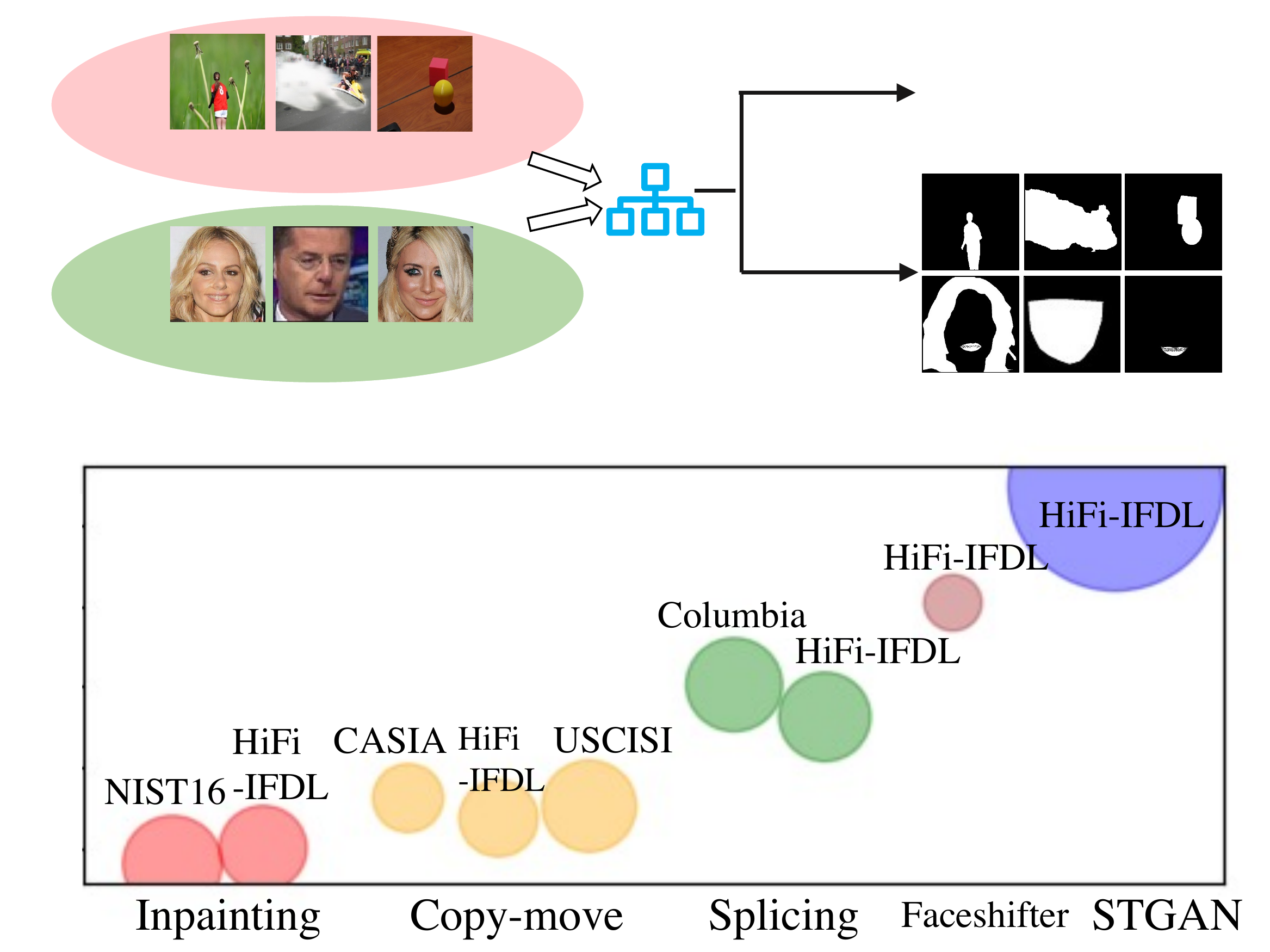}
    
    % upper figure
    % \put(48,45){\small{Model}}
    \put(57.5,69){\small{Detection}}
    \put(54,50.5){\small{Localization}}
    \put(73,66.5){\small{\textcolor{red}{Real} \textit{v.s.} Forgery}}
    \put(15,61.5){\small{Image Editing}}
    \put(12,46.5){\small{CNN-synthesized}}
    \put(49,41){(a)}
    
    % low figure.
    \put(0,1){\rotatebox{90}{\tbf{\scriptsize{Mean/Variance of forgery area}}}}
    \put(3.5,33){\footnotesize{$.5$}}
    \put(3.5,26.5){\footnotesize{$.4$}}
    \put(3.5,20){\footnotesize{$.3$}}
    \put(3.5,13.5){\footnotesize{$.2$}}
    \put(3.5,7.5){\footnotesize{$.1$}}
    % \put(7,3){\footnotesize{$.0$}}
    \put(49,-1.5){(b)}
    
    % \put(12,12){\scriptsize{\cite{NIST16}}}
    % \put(25,17.5){\scriptsize{\cite{dong2013casia}}}
    
    \end{overpic}
    \caption{(a) In this work, we study image forgery detection and localization (IFDL), regardless of forgery method domains. %[Keys: Det.: Detection; Loc.: Localization]. 
    (b) The distribution of forgery region depends on individual forgery methods. Each color represents one forgery category (x-axis). Each bubble represents one image forgery dataset. The y-axis denotes the average of forgery area. The bubble's area is proportional to the variance of the forgery area.}
    \label{fig_mani_overview}
    \vspace{-2mm}
\end{figure}
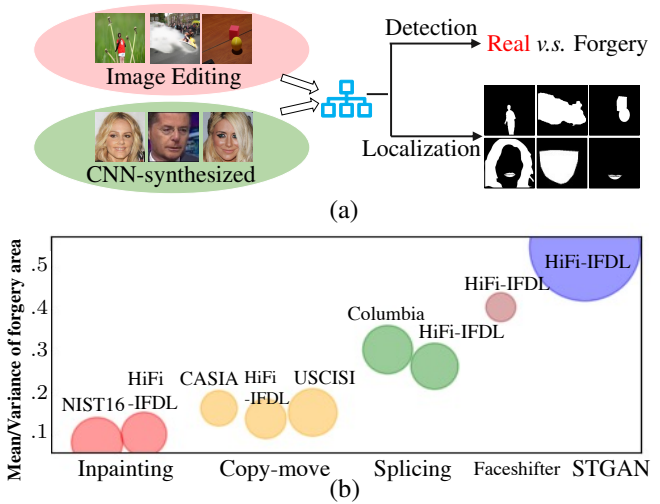

%% file: section/figure_table_latex/figure_overview.tex
\begin{figure*}[t]
    \centering
    \begin{overpic}[width=1\textwidth]{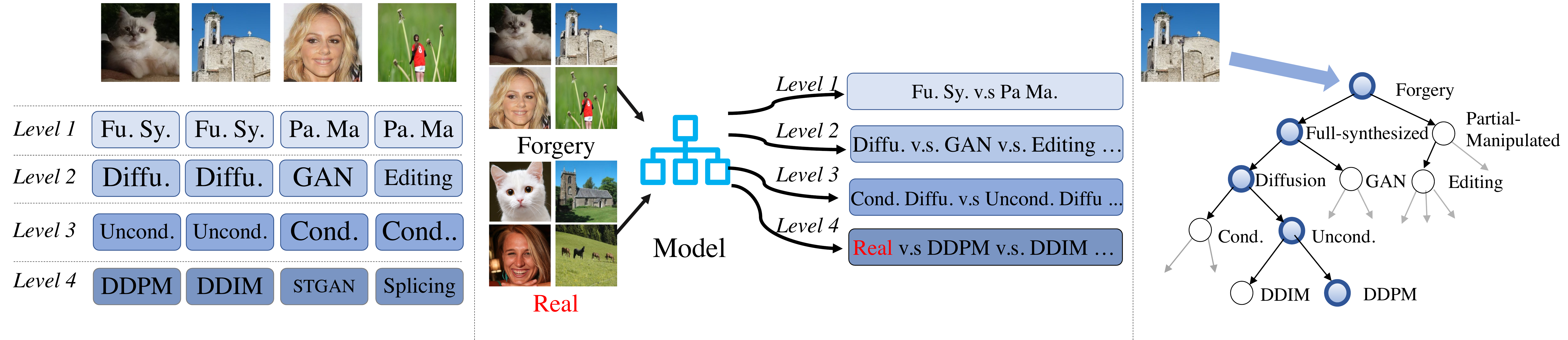}
    \put(18,0){(a)}
    \put(50,0){(b)}
    \put(85,0){(c)}
    % \put(76,21){\scriptsize{$Pr(\text{DDPM}) = Pr(\text{DDPM}\mid \text{Diffusion} )\times$}}
    % \put(82,19){\scriptsize{$Pr(\text{Diffusion}\mid \text{Ful. Syn.})\times$}}
    % \put(82,17){\scriptsize{$Pr(\text{Ful. Syn.}\mid \text{Forgery})\cdots$}}
    \end{overpic}
    \caption{(a) We represent the forgery attribute of each manipulated image with multiple labels, at different levels. (b) For an input image, we encourage the algorithm to classify its fine-grained forgery attributes at different levels, \textit{i.e.} a $2$-way classification (fully synthesized or partially manipulated) on level $1$. (c) We perform the fine-grained classification via the hierarchical nature of different forgery attributes, where each depth $l$ node's classification probability is conditioned on classification probabilities of neighbor nodes at depth ($l-1$). [Key: Fu. Sy.: Fully Synthesized; Pa. Ma.: Partially manipulated; Diff.: Diffusion model; Cond.: Conditional; Uncond.: Unconditional].}
    
    % \begin{overpic}[width=0.5\textwidth]{figure/overview_v32.pdf}
    % \end{overpic}
    % \begin{overpic}[width=0.48\textwidth]{figure/overview.pdf}
    % \put(32,0){\tiny\cite{liu2022pscc}}
    % \put(38,0){\tiny\cite{liu2022pscc}}
    % \put(44,0){\tiny\cite{wu2018busternet_copy_move_yue_wu}}
    % \put(50,0){\tiny\cite{li2019faceshifter}}
    % \put(56,0){\tiny\cite{liu2019stgan}}
    % \put(61,0){\tiny\cite{choi2020starganv2}}
    % \put(66,0){\tiny\cite{li2021image_hisd}}
    % \put(72,0){\tiny\cite{dhariwal2021diffusion_guided}}
    % \put(78,0){\tiny\cite{rombach2021highresolution_latent_diffusion}}
    % \put(84,4){$\cdots$ $\cdots$}
    % \put(90,10){$\cdots$}
    % \end{overpic}
    % \caption{\textit{Top}: We study the image manipulation detection and localization (IMDL), regardless of whether manipulation method is either image editing or CNN-manipulated. \textit{Bottom}: We develop a taxonomy fine-grain formulation which encourages the model to predict the entire taxonomy path to the certain generative method. Such formulation encourages model to learn the better representation for the IMDL and can help learn manipulation attribute. 
    % % \textit{Bottom}: To generalize towards images from unseen generative methods, the pre-trained model can parse input images attribute at different taxonomy levels.
    % }
    \vspace{-2mm}
    \label{fig_overview_2}
\end{figure*}

%% file: section/02_related_work.tex
\Section{Related Work}\label{sec:related}

\Paragraph{Image Forgery Detection} 
In the generic image forgery detection, it is required to distinguish real images from ones generated by a CNN: Zhang \etal \cite{zhangxue2019detecting} report that it is difficult for classifiers to generalize across different GANs and leverage upsampling artifacts as a strong discriminator for GAN detection. 
On the contrary, against expectation, the work by Wang \etal~\cite{wang2020cnn} shows that a baseline classifier \emph{can} actually generalize in detecting different GAN models contingent to being trained on synthesized images from ProGAN~\cite{karras2018progressive}. %with $20$ categories augmented with transformations.
% The proactive scheme~\cite{proactive-image-manipulation-detection} also improves generalization. 
Another thread is facial forgery detection~\cite{KorshunovICB2019,rossler2019faceforensics++v3,google_dfd,dolhansky_deepfake_2019,li_2020_CVPR,Jiang_2020_CVPR,unified-detection-of-digital-and-physical-face-attacks,proactive-image-manipulation-detection,bui2022repmix},
and its application in bio-metrics~\cite{guo2022multi,guo2019human,guo2020cord19sts,hsu2021discourse,abdalmageed2020assessment}. 
All these works specialize in the image-level forgery detection, which however does not meet the need of knowing where the forgery occurs on the pixel level. 
Therefore, we perform both image forgery detection and localization, as reported in Tab.~\ref{tab_background_compare}.

% Our work shares similar traits with prior work, such as localizing the manipulation and exploiting the hierarchy of manipulation as in ManTra-Net~\cite{wu2019mantra} yet, unlike ManTra-Net, ours does not perform classification with a linear classification layer directly yet uses a full hierarchical tree to achieve detection and classification. Additionally, 
% Unlike~\cite{wang2020cnn} our method provides spatial information for the manipulation and similarly to the GAN attribution task is able to indicate the most confident path in the hierarchy, refraining to reporting underconfident predictions at a lower level of the tree.
\input{section/figure_table_latex/table_comparison_related}
\Paragraph{Forgery Localization} 
% Localization of the manipulated portion of the image is of extreme importance for explainability of AI and can be considered the initial building block towards explainable AI for media forensics.
% Early work on media forensic focus on detecting a specific manipulation type, such as, splicing~\cite{uricchio2017localization,bondi2017tampering,cozzolino2015splicebuster,cozzolino2019noiseprint,huh2018fighting,kniaz2019point,lyu2014exposing,wu2017deep}, copy-move~\cite{bappy2019hybrid,islam2020doa,wen2016coverage,wu2018busternet_copy_move_yue_wu,wu2018image}, and, more recently,  inpainting~\cite{zhu2018deep}. Recently, ManTra-Net~\cite{wu2019mantra} was able to detect multiple manipulation types. Particularly, ManTra-Net treats manipulation traces as anomalous feature and adopts additional annotated data to optimize the model. SPAN~\cite{hu2020span} further proposes a pyramid model with local self-attention block that helps learn spatial correlation.
Most of existing methods perform pixel-wise classification 
% at the same resolution of the input image 
to identify forged regions~\cite{wu2019mantra,hu2020span,wang2022objectformer} while early ones use a region~\cite{zhou2018learning} or patch-based~\cite{mayer2018learned} approach. The idea of localizing forgery is also adopted in the DeepFake Detection community by segmenting the artifacts in facial images~\cite{zhao2021learning,chai2020makes,cozzolino2018forensictransfer}. 
% Stehouwer \etal~\cite{stehouwer2019detection} localize the region that contains man-made artifacts based on the standard $\ell_1$ loss and the whole architecture is optimized by minimizing the classification and localization losses. Similarly, Zhao \etal~\cite{zhao2020learning} use the binary cross-entropy (BCE) to supervise the $4$D volume employed in the self-consistency branch of their model. Recently, \cite{huang2020fakelocator} uses the face parsing as the auxiliary function to help localize the editing area.
%%% IAC Zhou et al is plural. It means Zhour and the others
Zhou \etal~\cite{zhou2020generate} improve the localization 
% by generating more true positive at training time and 
by focusing on object boundary artifacts. 
The MVSS-Net~\cite{chen2021image,dong2022mvss} uses multi-level supervision 
% at the pixel, edge, and image-level 
to balance between sensitivity and specificity.
% in the manipulation segmentation task. 
%ObjectFormer~\cite{wang2022objectformer} exploits the consistency at the object-level to improve localization. 
MaLP~\cite{malp-manipulation-localization-using-a-proactive-scheme} shows that the proactive scheme benefits both detection and localization. 
% It does so by using object prototypes as mid-level representations to model differences of region at the object level using attention mechanism. The output of the mid-level representation is further employed to refine patch embeddings to capture the patch-level consistencies. 
While prior methods are restricted to one domain, our method  unifies  across different domains.
% is not restricted to a single generative method domain and 
% makes an effort towards a well-rounded 
% for two forgery domains.
% regardless of the diverse spectrum of forgery types.

\Paragraph{Attribute Learning} CNN-synthesized image attributes can be observed in the frequency domain~\cite{zhangxue2019detecting,wang2020cnn}, where different GAN generation methods have distinct high-frequency patterns. 
The task of ``GAN discovery and attribution'' attempts to identify the exact generative model~\cite{marra2018detection,yu2019attributing_image_attribute,marra2019gans} while ``model parsing'' identifies both the model and the objective function~\cite{asnani2021reverse}. These works differ from ours in two aspects. 
Firstly, the prior work concentrates the attribute used in the digital synthesis method (synthesis-based), yet our work studies forgery-based attribute, \textit{i.e.}, to classify GAN-based fully-synthesized or partial manipulation from the image editing process. 
Secondly, unlike the prior work that assumes a ``flat'' structure between different attributes, we represent all forgery attributes in a hierarchical way, exploring dependencies among them.

%% file: section/figure_table_latex/table_comparison_related.tex
\begin{table}[t!]
\centering
\scriptsize
\resizebox{0.5\textwidth}{!}{
\begin{tabular}{c|c|c|c|c}
\hline
 Method & Det. & Loc. & \begin{tabular}[c]{@{}c@{}}Forgery\\Type\end{tabular} & \begin{tabular}[c]{@{}c@{}}Attribute\\Learning\end{tabular}\\\hline

Wu \etal~\cite{wu2019mantra} & \textcolor{lightred}{\ding{56}} & \textcolor{citecolor}{\ding{52}} & Editing & \textcolor{lightred}{\ding{56}}\\ 

Hu \etal~\cite{hu2020span} & \textcolor{lightred}{\ding{56}} & \textcolor{citecolor}{\ding{52}} & Editing & \textcolor{lightred}{\ding{56}}\\ 

Liu \etal~\cite{liu2022pscc} & \textcolor{citecolor}{\ding{52}} & \textcolor{citecolor}{\ding{52}} & Editing & \textcolor{lightred}{\ding{56}}\\ 

Dong \etal~\cite{dong2022mvss} & \textcolor{citecolor}{\ding{52}} & \textcolor{citecolor}{\ding{52}} & Editing & \textcolor{lightred}{\ding{56}}\\ 

Wang \etal~\cite{wang2022objectformer} & \textcolor{citecolor}{\ding{52}} & \textcolor{citecolor}{\ding{52}} & Editing & \textcolor{lightred}{\ding{56}}\\ 

% \cite{mahfoudi2019defacto} & \textcolor{citecolor}{\ding{52}} & \textcolor{lightred}{\ding{56}} & A.I. \& Editing & \textcolor{lightred}{\ding{56}} & $2019$\\ 

% x & \checkmark & x & x
Zhang \etal~\cite{zhangxue2019detecting} & \textcolor{citecolor}{\ding{52}} & \textcolor{lightred}{\ding{56}} & CNN-based & \textcolor{lightred}{\ding{56}}\\

Wang \etal~\cite{wang2020cnn} & \textcolor{citecolor}{\ding{52}}& \textcolor{lightred}{\ding{56}} & CNN-based & \textcolor{lightred}{\ding{56}}\\ 

%%%%%%%%%%%%%%%%%%%%%%%%%%%%%%%%%%%%%%%%%%%%%
Asnani \etal~\cite{asnani2021reverse} & \textcolor{citecolor}{\ding{52}} & \textcolor{lightred}{\ding{56}} & CNN-based & syn.-based\\

Yu \etal~\cite{yu2019attributing_image_attribute} & \textcolor{citecolor}{\ding{52}} & \textcolor{lightred}{\ding{56}} & CNN-based & syn.-based\\
% & x & \checkmark & x & x

%%%%%%%%%%%%%%%%%%%%%%%%%%%%%%%%%%%%%%%%%%%%%
Stehouwer \etal~\cite{stehouwer2019detection} & \textcolor{citecolor}{\ding{52}} & \textcolor{citecolor}{\ding{52}} & CNN-based & \textcolor{lightred}{\ding{56}}\\

Huang \etal~\cite{huang2020fakelocator} & 
\textcolor{citecolor}{\ding{52}} & \textcolor{citecolor}{\ding{52}} & CNN-based & \textcolor{lightred}{\ding{56}}\\

% \hline
\hline
Ours & \textcolor{citecolor}{\ding{52}} & \textcolor{citecolor}{\ding{52}} & Both types & for.-based\\
\hline
\end{tabular}
}
\vspace{-2mm}
\caption{Comparison to previous works. [Key: Det.: detection, Loc.: localization, Syn.: synthesis, for.: forgery]  \vspace{-3mm}}
\label{tab_background_compare}
\end{table}

% \begin{table}[t]
%     \centering 
%     \resizebox{1.\linewidth}{!}{
%     {\def\arraystretch{1.2}
%      \begin{tabular}{@{}l|ccccr@{}}
%     \toprule Dataset & \shortstack[l]{Img Editing \\ detection} & \shortstack[l]{A.I. Generated\\ detection} & \shortstack[l]{Manip.\\ Attribution} & \shortstack[l]{Manip.\\Localization} \\
%     \midrule
%     DEFACTO~\cite{mahfoudi2019defacto}  & \checkmark & x & x & \checkmark\\%\cdashline{1-8}
%     DFDD~\cite{stehouwer2019detection}  & x & \checkmark & x & \checkmark\\%\cdashline{1-8}
%     GAN-fingerprint~\cite{yu2019attributing_image_attribute}&x&x&\checkmark& x\\
%     Reverse~\cite{asnani2021reverse} & x & \checkmark & x & x \\
%     AutoGAN~\cite{zhang2019detecting}  & x & \checkmark & x & x\\
%     ForenSynths~\cite{zhang2019detecting}  & x & \checkmark & x & x \\%\cdashline{1-8}
%     ForgeryNet~\cite{he2021forgerynet}  & x & \checkmark & x & \checkmark \\
%     \cdashline{1-6}
%     Ours  & \red{\textbf{\checkmark}} & \red{\textbf{\checkmark}} & \red{\textbf{\checkmark}} & \red{\textbf{\checkmark}}\\
%     \bottomrule
%     \end{tabular}}
%     }
%     \caption{The comparison to the prior works. Our manipulation detection and localization work covers both A.I. generated image and conventional image editing. Also, our method can be used for model parsing.}
%     \label{tab:datastats}
%     \vspace{-5mm}
% \end{table} 

%% file: section/03_00_model.tex
\input{section/figure_table_latex/figure_architecture}
\Section{HiFi-Net}
\label{sec:method}
In this section, we introduce HiFi-Net as shown in Fig.~\ref{fig:architecture}. We first define the image forgery detection and localization (IFDL) task and hierarchical fine-grained formulation. 
In IFDL, an image $\mbf{X} \in \takeval{R}{0}{255}^{3 \PLH W \PLH H}$ is mapped to a binary variable $\mbf{y}$
% categorical variable $\mbf{Y}$ 
for image-level forgery detection
% multi-class classification, 
and a binary mask $\mbf{M} \in \takeval{R}{0}{1}^{W \PLH H}$ for localization, where the $\mbf{M}_{ij}$ indicates if the $ij$-th pixel is manipulated or not. 
%%% Dear Iacopo:
%%% So the IFDL tasks produce a binary label for each image, but we train it with fine-grained hierarchical formulation, which encourage the algorithm to do fine-grained classification.

In the hierarchical fine-grained formulation, we train the given IFDL algorithm towards fine-grained classifications, and in the inference we evaluate the binary classification results on the image-level forgery detection. Specifically, we denote a categorical variable $\hat{\mbf{y}}_b$ at branch $b$, where its value depends on which level we conduct the fine-grained forgery attribute classification. 
For example, as depicted in Fig.~\ref{fig_overview_2}\textcolor{red}{b}, two categories at level $1$ are full-synthesized, partial-manipulated; four classes at level $2$ are diffusion model, GAN-based method, image editing, CNN-based partial-manipulated method; classes at level $3$ discriminate whether forgery methods are conditional or unconditional; $14$ classes at level $4$ are real and $13$ specific forgery methods.
We detail this in Sec.~\ref{sec_benchmark} and Fig.~\ref{fig_taxonomy_benchmark}. 

To this end, we propose HiFi-Net (Fig.~\ref{fig:architecture}) which consists of a multi-branch feature extractor (Sec.~\ref{sec_multi_scale}) that performs fine-grained classifications at different specific forgery attribute levels, and two modules (Sec.~\ref{sec_loc} and Sec.~\ref{sec_cls}) that help the forgery localization and detection, respectively. Lastly, Sec.~\ref{sec_train_infer} introduces training procedure and inference.

\SubSection{Multi-Branch Feature Extractor} \label{sec_multi_scale}
We first extract feature of the given input image via the color and frequency blocks, and this frequency block applies a Laplacian of Gaussian (LoG)~\cite{burt1987laplacian_laplacia_operator} onto the CNN feature map. This architecture design is similar to the method in~\cite{masi2020two}, which exploits image generation artifacts that can exist in both RGB and frequency domain~\cite{wang2022objectformer,dong2022mvss,wang2020cnn,zhangxue2019detecting}.

Then, we propose a multi-branch feature extractor, and whose branch is denoted as $\net_b$ with $ b \in \{1 \ldots 4\}$. Specifically, each $\net_b$ generates the feature map of a specific resolution, and such a feature map helps $\net_b$ conduct the fine-grained classification at the corresponding level. For example, for the finest level (\textit{i.e.}~identifying the individual forgery methods), one needs to model contents at all spatial locations, which requires high-resolution feature map.
In contrast, it is reasonable to have low resolution feature maps for the coarsest level (\textit{i.e.}~binary) classification. 
% Thus, we construct a multi-branch feature extractor where each branch operates on feature maps at a specific resolution and performs classification at a specific level.

We observe that different forgery methods generate manipulated areas with different distributions (Fig.~\ref{fig_mani_overview}\textcolor{red}{b}), and different patterns, \textit{e.g.}, deepfake methods~\cite{rossler2019faceforensics++v3,li2019faceshifter} manipulate the whole inner part of the face, whereas STGAN~\cite{liu2019stgan} changes sparse facial attributes such as mouth and eyes.
%In image editing, splicing involves large area, yet inpainting can only erase objects of the small size. % We notice the localization performance is associated with the area of manipulation region, so we place the localization module at the branch that is of the highest resolution. 
Therefore, we place the localization module at the end of the highest-resolution  branch of the extractor---the branch to classify specific forgery methods. 
In this way, features for fine-grained classification serve as a prior for localization. 
It is important to have such a design for localizing both manipulated images with CNNs or classic image editing.

\SubSection{Localization Module}
\label{sec_loc}
\Paragraph{Architecture} 
%The model exposes a localization module that is optimized to match the ground-truth binary mask $\mbf{M}$. 
% The localization module branches from the last convolutional feature map associated with the highest-resolution branch and is optimized to match the ground-truth binary mask $\mbf{M}$ associated with the input image.
The localization module maps feature output from the highest-resolution branch ($\net_{4}$), denoted as $\mathbf{F}\in \mathbb{R}^{512\PLH W \PLH H}$, to the mask $\hat{\mbf{M}}$ to localize the forgery. To model the dependency and interactions of pixels on the large spatial area, the localization module employs the self-attention mechanism~\cite{zhang2019self_att_gan,wang2018non_local}. 
As shown in the localization module architecture in~\cref{fig_localization_module}, 
% In the localization module, 
we use $1\PLH 1$ convolution to form $g$, $\phi$ and $\psi$, which convert input feature $\mathbf{F}$ into $\mathbf{F}_{g} = g(\mathbf{\mathbf{F}})$, $\mathbf{F}_{\phi} = \phi(\mathbf{\mathbf{F}})$ and $\mathbf{F}_{\psi} = \psi(\mathbf{\mathbf{F}})$. Given $\mathbf{F}_{\phi}$ and $\mathbf{F}_{\theta}$, we compute the spatial attention matrix $\mathbf{A}_{s} = \operatorname{softmax} (\mathbf{F}_{\phi}^{T} \mathbf{F}_{\theta})$. We then use this transformation $\mathbf{A}_{s}$ to map $\mathbf{F}_{g}$ into a global feature map  $\mbf{F}^{\prime}=\mathbf{A}_{s} \mathbf{F}_{g} \in \mathbb{R}^{512\PLH W \PLH H}$.

\Paragraph{Objective Function} Following~\cite{masi2020two}, we employ a metric learning objective function for localization, which creates a wider margin between real and manipulated pixels.
% and thus may generalize well to unseen manipulation domains.
% The motivation is this objective function has better separation and can generalize to different unseen real domain data. 
%Specifically, we transform $\mbf{F}_{\text{img}}$ into a binary mask $\mbf{M} \in \takeval{R}{0}{1}^{W \PLH H}$ that indicates the likelihood of being manipulated at each spatial locations.
% Unlike the competitive approach~\cite{wu2019mantra} that uses softmax regression as the classifier, 
We firstly learn features of each pixel, and then model the geometry of such learned features with a radial decision boundary in the hyper-sphere. Specifically, we start with pre-computing a reference center $\mathbf{c} \in \mathbb{R}^{D}$, by averaging the features of all pixels in real images of the training set.
% and $D$=$18$.
%%%%%%%%%%%%%%%%%%%%%%%%%%%%%%%%%
% \red{Better to use small $\mbf{x}$ and not $\mbf{X}$ }
We use $\mbf{F}^{\prime}_{ij}\in \mathbb{R}^{D}$ to indicate the $ij$-th pixel of the final mask prediction layer. Therefore, our localization loss $\mathcal{L}_{loc}$ is:
\vspace{-2mm}
\begin{equation}
\small
\mathcal{L}_{loc} = \frac{1}{HW} \sum_i^H\sum_j^W \mathcal{L}\big(\mbf{F}^{\prime}_{ij}, \mbf{M}_{ij}; \mathbf{c}, \tau\big),
\vspace{-3mm}
\label{eq_L_loc}
\end{equation}
where:
\begin{equation*}
\small
\mathcal{L} = \begin{cases} \left \| \mbf{F}^{\prime}_{ij} - \mathbf{c} \right \|_2 & \mbox{if } 
 \mbf{M}_{ij} \mbox{ real} \\ 
 \max\big(0, \tau - \left \| \mbf{F}^{\prime}_{ij} - \mathbf{c} \right \|_2\big) & \mbox{if } \mbf{M}_{ij}\mbox{ forged}. \end{cases}
\label{eq_deep_loss}
\end{equation*}
Here $\tau$ is a pre-defined margin. 
% set to $2.5$ empirically. ==> Dr. Liu says this should be in the implementation details.
The first term in $\mathcal{L}$ improves the feature space compactness of real pixels. 
The second term encourages the distribution of forged pixels to be far away from real by a margin $\tau$. 
Note our method differs to \cite{ruff2018deep,masi2020two} in two aspects: 1) unlike~\cite{ruff2018deep}, we use the second term in $\mathcal{L}$ to enforce separation; 2) compared to the image-level loss in~\cite{masi2020two} that has two margins, we work on the more challenging pixel-level learning. Thus we use a single margin, which reduces the number of hyper-parameters and improve the simplicity.

\SubSection{Classification Module}
\label{sec_cls}
\Paragraph{Partial Convolution} 
% We intend to reuse the generated localization map, for achieving the best representation learning which finds the optimal solution to both localization and classification. This is different to the previous work whose target is merely to localize the manipulation region with a binary map or heat map.
Unlike prior work~\cite{wang2022objectformer,dong2022mvss,hu2020span} whose ultimate goal is to localize the forgery mask, we reuse the forgery mask to help HiFi-Net learn the optimal feature for classifying fine-grained forged attributes. Specifically, we generate a binary mask $\hat{\mathbf{M}}$, then overlay $\hat{\mathbf{M}}$ with the input image as $\mathbf{X} \odot \hat{\mathbf{M}}$ to obtain the masked image $\mathbf{X}_{mask} \in \mathbb{R}^{3 \PLH W_{0} \PLH H_{0}}$. 
To process the  masked image, we resort to the \emph{partial convolution} operator (PConv)~\cite{liu2018image_partialcnn}, whose convolution kernel is renormalized to be applied only on unmasked pixels.
% as depicted in Fig~\ref{fig_localization_module}.
% specified by the mask. 
The idea is to have feature maps only describe pixels at the manipulated region.
% and use the mask to exclude the others. 
PConv acts as conditioned dot product for each kernel, conditioned on the mask. Denoting $\mathbf{W}_{par}$ as the convolution kernel, we have:
\begin{equation}
    \small
    \mathbf{X}^{\prime}_{mask} =
        \mathbf{W}_{par}^T \mathbf{X}_{mask} = \mathbf{W}_{par}^T (\mathbf{X} \odot \hat{\mathbf{M}}),
\label{eq_pconv}
\end{equation}
where the dot product $\odot$ is ``renormalized'' to account for zeros in the mask. At different layers, we update and propagate the new mask $\hat{\mathbf{M}}^{\prime}$ according to the following equation:
\begin{equation}
    \small
    \hat{\mathbf{M}}^{\prime} = 
    \begin{cases}
        1 & \text{If      }\| \hat{\mathbf{M}} \| \ge 0 \\
        0 & \text{otherwise}. \\
    \end{cases}
\end{equation}

Specifically, $\mathbf{X}_{mask}$ represents the most prominent forged image region. 
We believe the feature of $\mathbf{X}_{mask}$ can serve as a prior for HiFi-Net, to better learn the attribute of individual forgery methods. 
For example, the observation whether the forgery occurs on the eyebrow or entire face, helps decide whether given images are manipulated by STGAN~\cite{liu2019stgan} or FaceShifter~\cite{li2019faceshifter}.
The localization part is implemented with only two light-weight partial convolutional layers for higher efficiency. 
\input{section/figure_table_latex/figure_localization}
\Paragraph{Hierarchical Path Prediction}
We intend to learn the hierarchical dependency between different forgery attributes. Given the image $\mbf{X}$, we denote output logits and predicted probability of the branch $\net_{b}$ as $\net_b(\mbf{X})$ and $p(\mbf{y}_b|\mbf{X})$, respectively. Then, we have: 
\begin{equation}
    \small
    \begin{split}
    p(\mbf{y}_b|\mbf{X}) \doteq \operatorname{softmax}\Big(\net_b(\mbf{X})\odot (1 + p(\mbf{y}_{b-1}|\mbf{X}))\Big) \quad
    \end{split}
\label{eq_conditional_prob}
\end{equation}
% Eq.~(\ref{eq_conditional_prob}) shows that to compute the conditional probability at branch $b$, which predicts fine-grained classes, the output logits are multiplied by the probabilities obtained by the branch $b-1$ at the previous coarser level.
Before computing the probability $p(\mbf{y}_b|\mbf{X})$ at branch $\net_{b}$, we scale logits $\net_b(\mbf{X})$ based on the previous branch probability $p(\mbf{y}_{b-1}|\mbf{X})$. 
% functioning as an attention mechanism.
% After that, predicted class probability on each node at level~$l$ is conditioned on class probability of its neighbor node at level~$(l-1)$.
%% IAC: we have to add this aotherwise the equation does not make sense
%% since the classes are no aligned across layers, by repeating you align them.
% Given that the number of classes between layers are different, we align the vectors 
Then, we enforce the algorithm to learn hierarchical dependency. Specifically, in Eq.~(\ref{eq_conditional_prob}), we repeat the probability of the coarse level $b-1$ for all the logits output by branch at level $b$, following the hierarchical structure. Fig.~\ref{fig_path_prediction} shows that the logits associated to predicting \texttt{DDPM} or \texttt{DDIM} are multiplied by probability for the image to be \texttt{Unconditional (Diffusion)} in the last level, according to the hierarchical tree structure.
\input{section/figure_table_latex/figure_tax_path_prediction}
\input{section/figure_table_latex/figure_IMDL_dataset}

\SubSection{Training and Inference}\label{sec_train_infer}
% In training, we use $\mathcal{L}_{loc}$ (Eq.~(\ref{eq_L_loc})) and the classification loss $\mathcal{L}_{cls}$. As 
In the training, each branch is optimized towards the classification at the corresponding level, we use $4$ classification losses, $\mathcal{L}^{1}_{cls}$, $\mathcal{L}^{2}_{cls}$, $\mathcal{L}^{3}_{cls}$ and $\mathcal{L}^{4}_{cls}$ for $4$ branches. At the branch $b$, $\mathcal{L}^{b}_{cls}$ is the cross entropy distance between $p(\mbf{y}_b|\mbf{X})$ and a ground truth categorical $\hat{\mbf{y}}_b$.
% Thus, $\mathcal{L}_{cls}$ is:
% \begin{equation}
%     \vspace{-1.5mm}
%     \mathcal{L}_{cls} = \mathcal{L}^{1}_{cls} + \mathcal{L}^{2}_{cls} + \mathcal{L}^{3}_{cls} + \mathcal{L}^{4}_{cls}.
%     \vspace{-1mm}
% \end{equation}
The architecture is trained end-to-end with different learning rates per layers. The detailed objective function is:
\begin{equation*}
\small
\mathcal{L}_{tot} = 
    \begin{cases} 
        \lambda \mathcal{L}_{loc} + \mathcal{L}^{1}_{cls} + \mathcal{L}^{2}_{cls} + \mathcal{L}^{3}_{cls} + \mathcal{L}^{4}_{cls} & \mbox{if $\mbf{X}$ is forged}\\ 
        
        \lambda \mathcal{L}_{loc} + \mathcal{L}^{4}_{cls} & \mbox{if $\mbf{X}$ is real}
    \end{cases}
\end{equation*}
where $\mbf{X}$ is the input image. When the input image is labeled as ``real'', we only apply the last branch ($\net_{4}$) loss function, otherwise we use all the branches. $\lambda$ is the hyper-parameter that keeps $\mathcal{L}_{loc}$ on a reasonable magnitude.
% and $\mathcal{L}_{loc}$ on the reasonable magnitude.
% The final objective function $\mathcal{L}$ for HiFi-Net then becomes,
% \begin{equation}
%     \vspace{-1.5mm}
%     \mathcal{L}_{tot} = \mathcal{L}_{cls} + \lambda \mathcal{L}_{loc},
%     \vspace{-1mm}
% \end{equation}
% where $\lambda$ keeps $\mathcal{L}_{cls}$ and $\mathcal{L}_{loc}$ on the reasonable magnitude.

In the inference, HiFi-Net generates the forgery mask from the localization module, and predicts forgery attributes at different levels. 
We use the output probabilities at level $4$ for forgery attribute classification. 
For binary ``forged vs.~real'' classification, we predict as forged if the highest probability falls in any manipulation method at level $4$.

%% file: section/figure_table_latex/figure_architecture.tex
\begin{figure*}[t]
  \centering
    % \begin{overpic}[width=0.95\textwidth]{figure/architecture.pdf}
    \begin{overpic}[width=0.96\textwidth]{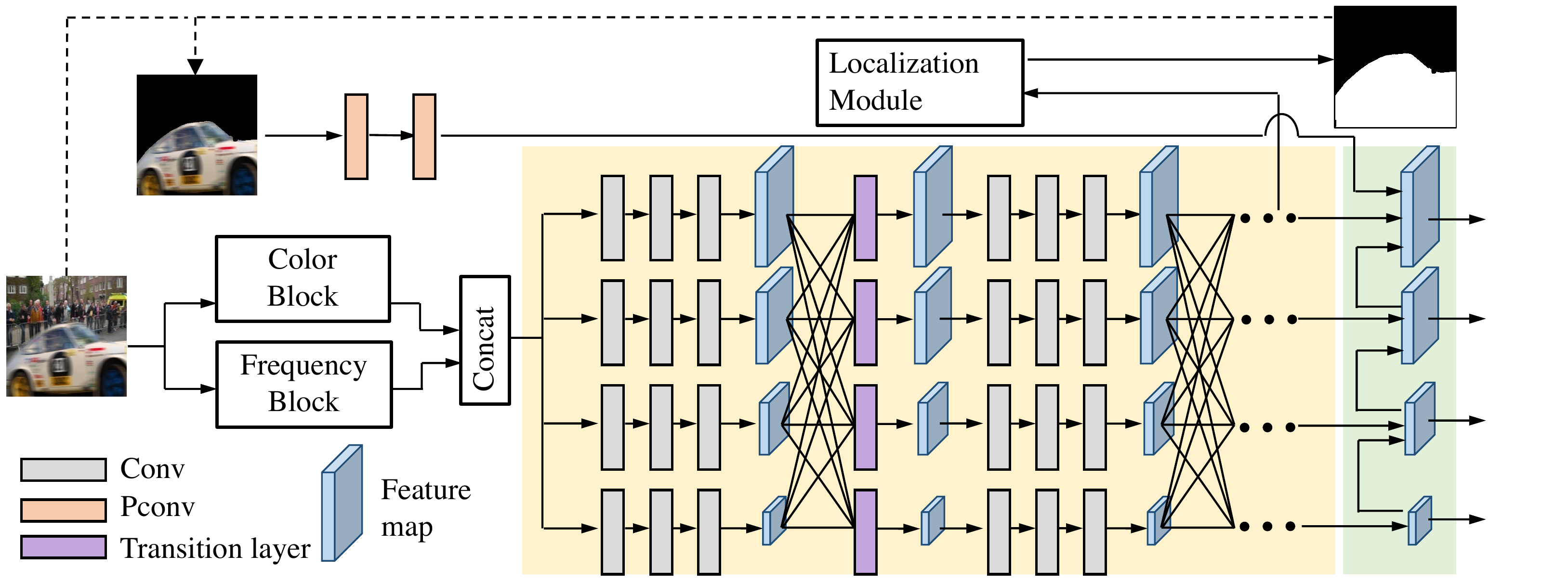}
    % architecture
    % \put(32,21.5){$\net_1$}
    % \put(32,13.5){$\net_2$}
    % \put(32,7.5){$\net_3$}
    % \put(32,3.5){$\net_4$}
    \put(1.5,10){Image}
    % \put(1,30){\small{PConv}}
    % \put(5.5,29.5){\footnotesize{\begin{tabular}{c}LoG\\Operator \end{tabular}}}
    % \put(14,30){\footnotesize{\begin{tabular}{c}Feature Extractor\\Branch\end{tabular}}}
    % \put(11,13.5){\footnotesize{Frequency Branch}}
    % \put(12,2){\footnotesize{Color Branch}}
    % \put(70,30){\footnotesize{\begin{tabular}{c}Localization\\Module\end{tabular}}}
    % classification module
    % \put(86,27){\footnotesize{\begin{tabular}{c}Classification\\Module\end{tabular}}}
    \put(94,31.5){\small{$\mathbf{\hat{M}}$}}
    \put(91,22.5){\scriptsize{\begin{tabular}{c}Real \textit{v.s.}\\Level~$4$\\ Forgery Attributes\end{tabular}}}
    \put(91,15.5){\scriptsize{\begin{tabular}{c}Level~$3$\\ Forgery Attributes\end{tabular}}}
    \put(91,9.3){\scriptsize{\begin{tabular}{c}Level~$2$\\ Forgery Attributes\end{tabular}}}
    \put(91,3){\scriptsize{\begin{tabular}{c}Level~$1$\\ Forgery Attributes\end{tabular}}}
    \end{overpic}
    \vspace{-1mm}
    \caption{Given the input image, we first leverage color and frequency blocks to extract features. The multi-branch feature extractor~(\textcolor{cell_blond}{\rule{0.4cm}{0.25cm}}) learns feature maps of different resolutions, for the fine-grained classification at different levels. The localization module (Sec.~\ref{sec_loc}) generates the forgery mask, $\hat{\mbf{M}}$, to identify the manipulation region. After that, we use the partial convolution (PConv) layer to encode the masked image (Eq.~\ref{eq_pconv}), and then leverage such ``masked'' embeddings in the classification module (\textcolor{mossgreen}{\rule{0.4cm}{0.25cm}}), which details in Sec.~\ref{sec_cls}.}
    \label{fig:architecture}
\end{figure*}
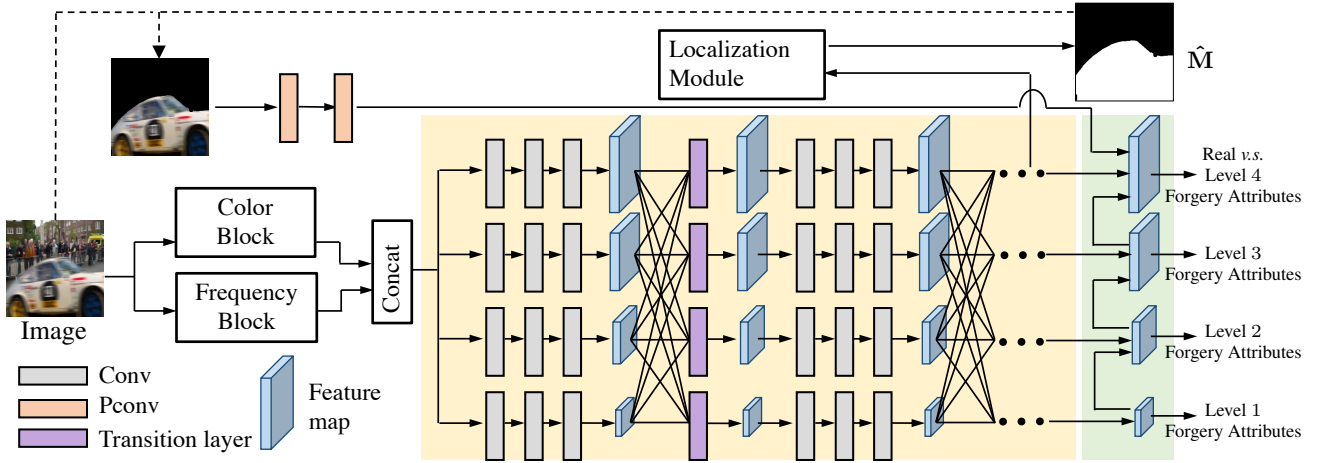

%% file: section/figure_table_latex/figure_localization.tex
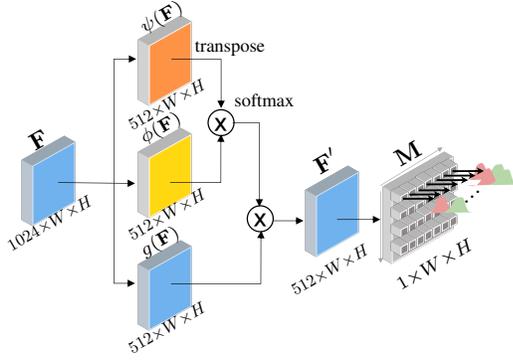
\begin{figure}[t]
    \centering
    % \includegraphics[width=\textwidth,height=5cm]{figure/partial_conv_fig.png}
    % \begin{overpic}[width=\textwidth]{figure/attention_metric_loss}
    \begin{overpic}[width=0.4\textwidth]{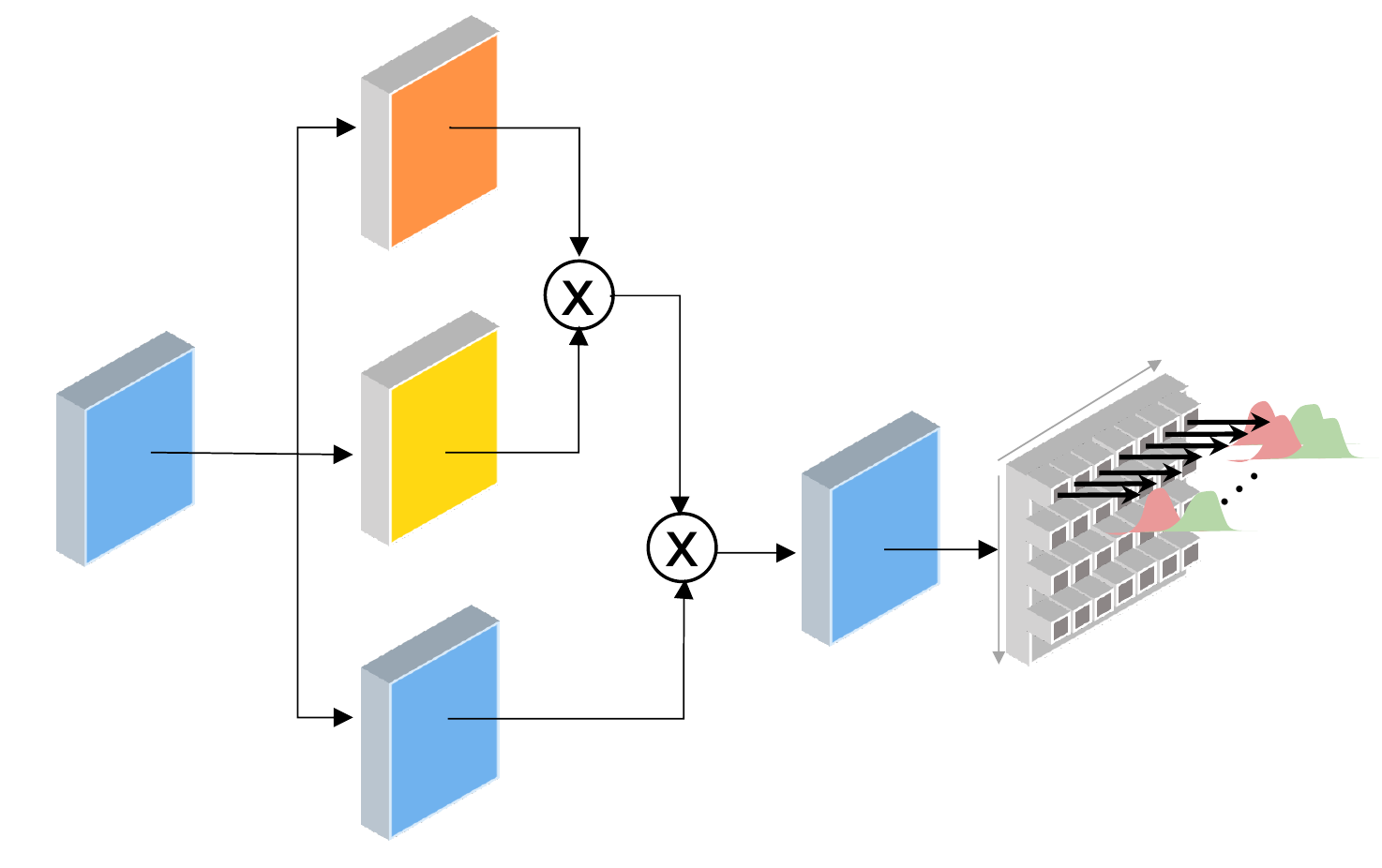}
    \put(5,36){\rotatebox{30}{\small{$\mathbf{F}$}}}
    \put(1,17){\scalebox{.7}{\rotatebox{30}{\small{$1024\PLH W \PLH H$}}}}
    
    \put(26,58.5){\rotatebox{30}{\scriptsize{$\psi(\mathbf{F})$}}}
    \put(24,40){\scalebox{.7}{\rotatebox{30}{\small{$512\PLH W \PLH H$}}}}
    \put(26,37.5){\rotatebox{30}{\scriptsize{$\mathbf{\phi(F)}$}}}
    \put(24,19){\scalebox{.7}{\rotatebox{30}{\small{$512\PLH W \PLH H$}}}}
    \put(26,16){\rotatebox{30}{\scriptsize{$g(\mathbf{F})$}}}
    \put(24,-2.5){\scalebox{.7}{\rotatebox{30}{\small{$512\PLH W \PLH H$}}}}
    
    % \put(18.45,20.65){\LARGE{${\times}$}}
    % \put(22.15,11.5){\LARGE{${\times}$}}
    
    \put(37,55){\scriptsize{transpose}}
    \put(44.5,44.5){\scriptsize{softmax}}
    
    \put(59,31.5){\rotatebox{30}{\small{$\mathbf{F^\prime}$}}}
    \put(55,8.5){\scalebox{.7}{\rotatebox{30}{\small{$512\PLH W \PLH H$}}}}
    
    % \put(33.5,17.5){\rotatebox{30}{\scriptsize{Mask Prediction Layer}}}
    \put(74.5,33){\rotatebox{30}{\small{$\mathbf{M}$}}}
    \put(75,9.5){\scalebox{1}{\rotatebox{30}{\scriptsize{$1\PLH W \PLH H$}}}}

    % pconvolution
    % \put(85,25){\small{$\mathbf{W}_{par.}$}}
    % \put(53,25){\small{$\mathbf{I}_{mask}$}}

    % % caption
    % \put(25,0){(a)}
    % \put(50,0){(b)}
    % \put(75,0){(c)}

    \end{overpic}
    \caption{The localization module adopts the self-attention mechanism to transfer the feature map $\mathbf{F}$ to the localization mask $\mathbf{M}$. 
    % (b) The PConv filter $\mathbf{W}_{par.}$ only works on the valid pixels region. (c) In the taxonomy path prediction, the classification probability at level~$l$ is conditioned on that at level~$(l-1)$, according to the pre-defined taxonomy tree.
    }\label{fig_localization_module}
    % \vspace{-3mm}
\end{figure}

% %%% Hyper-sphere from now on
% \put(65,12.5){$\mbf{c}$}
% \put(90,13){$\mbf{c}$}
% \put(60,10){\tiny{$\left \| \Phi(x_i) - \mbf{c} \right \|$}}
% \put(88.5,5){\tiny{Uncertainty region}} 
% \put(88.5,4){\tiny{is masked out}}
% \put(63.7,3.7){\tiny{Manipulate: pushed away}}
% \put(63.7,2.3){\tiny{Authentic: pulled to $\mbf{c}$ }}
% \put(63.7,0.75){\tiny{No force applied}}
% %\put(10,27){$\boldsymbol{\phi}(\mathbf{x})$}

%% file: section/figure_table_latex/figure_tax_path_prediction.tex
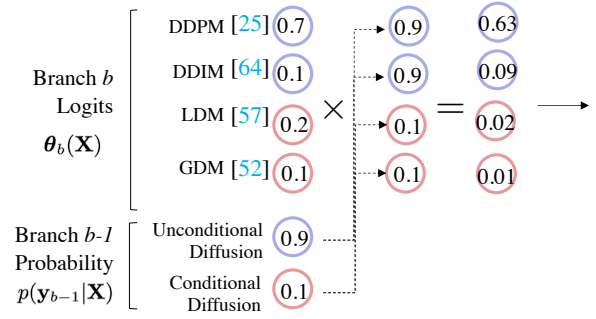
\begin{figure}[!t]
    \centering
    \begin{overpic}[width=0.95\linewidth]{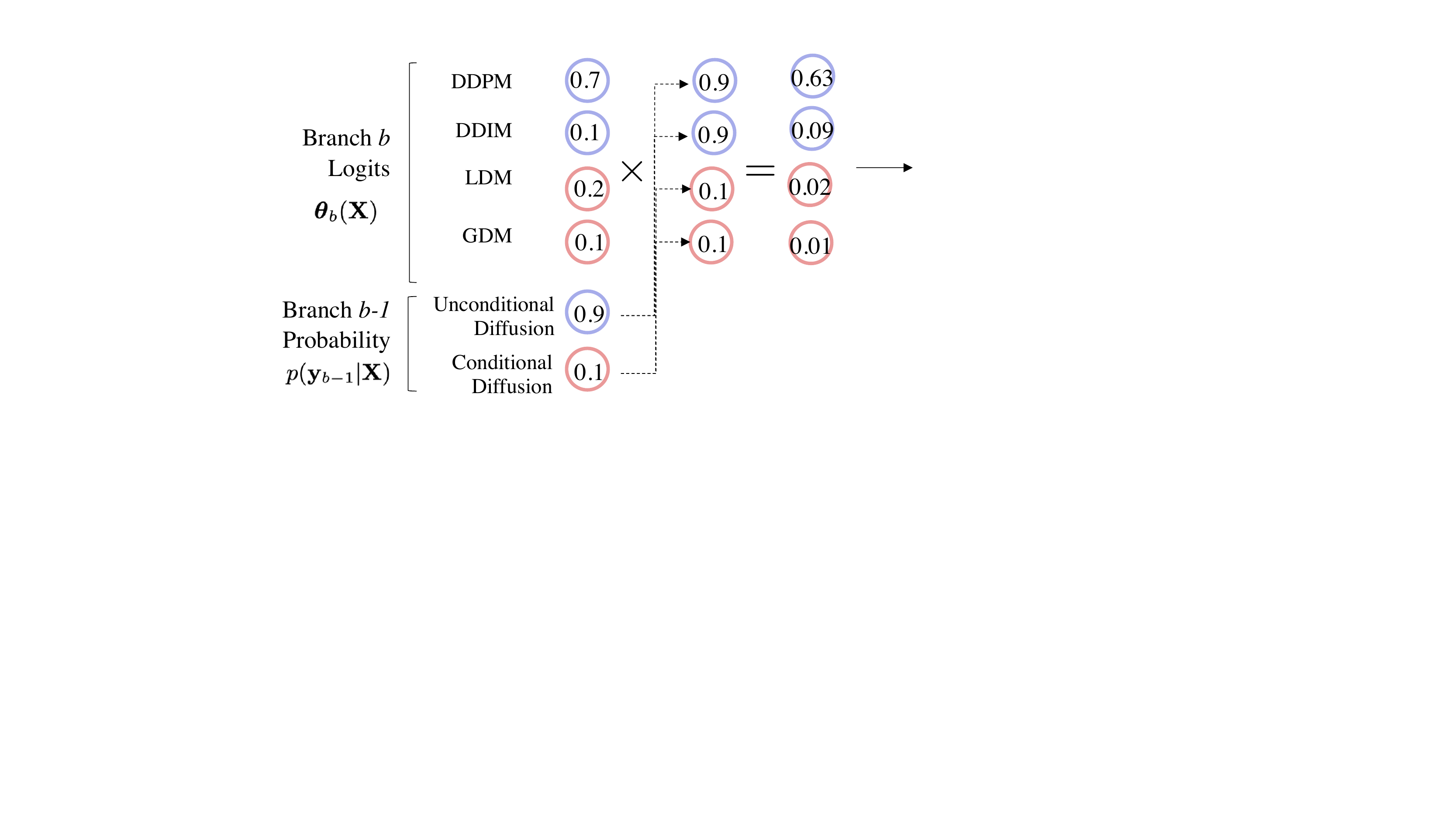}
    % \put(87,39){\small{Softmax}}
    \put(37.5,48.5){\small{\cite{ho2020denoising_ddpm}}}
    \put(37.5,41){\small{\cite{song2020denoising_ddim}}}
    \put(37.5,33){\small{\cite{rombach2021highresolution_latent_diffusion}}}
    \put(37.5,24){\small{\cite{nichol2021glide}}}
    % pconvolution
    % \put(85,25){\small{$\mathbf{W}_{par.}$}}
    % \put(53,25){\small{$\mathbf{I}_{mask}$}}

    % % caption
    % \put(25,0){(a)}
    % \put(50,0){(b)}
    % \put(75,0){(c)}
    \end{overpic}
    \caption{The classification probability output from branch $\net_{b}$ depends on the predicted probability at branch $\net_{b-1}$, following the definition of the hierarchical forgery attributes tree.}\label{fig_path_prediction}
    \vspace{-3mm}
\end{figure}

%% file: section/figure_table_latex/figure_IMDL_dataset.tex
\begin{figure*}[t]
    \centering
    \begin{subfigure}[b]{0.5\linewidth}
        \begin{overpic}[width=0.95\linewidth]{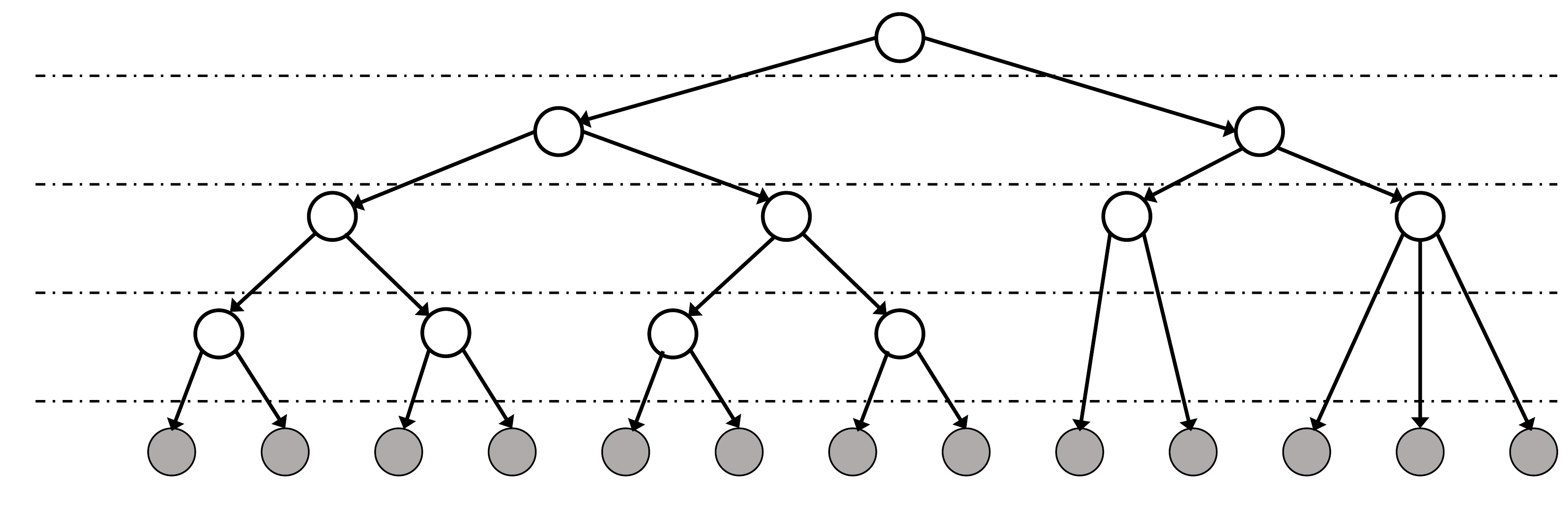}
        % levels
        \put(-3,25){\small{Level $1$}}
        \put(-3,17){\small{Level $2$}}
        \put(-3,11){\small{Level $3$}}
        \put(-3,3){\small{Level $4$}}
        
        % level 0,1 
        \put(63,31){\scriptsize{Forgery}}
        \put(42,25){\scriptsize{Fully-synthesized}}
        \put(84,25){\scriptsize{Partial-}}
        \put(84,23){\scriptsize{manipulated}}
        
        % level 2
        \put(26,18.5){\scriptsize{Diffusion}}
        \put(54,18.5){\scriptsize{GAN}}
        \put(76,18.5){\scriptsize{CNN-}}
        \put(76,16){\scriptsize{based}}
        \put(94,18.5){\scriptsize{Image}}
        \put(94,16){\scriptsize{editing}}
        
        % level 3
        \put(16,10){\scriptsize{Uncond..}}
        \put(31,10){\scriptsize{Cond.}}
        \put(45,10){\scriptsize{Uncond.}}
        \put(60,10){\scriptsize{Cond.}}
        
        % level 4
        \put(8,0){\scriptsize{\cite{nichol2021glide}}}
        \put(15,0){\scriptsize{\cite{rombach2021highresolution_latent_diffusion}}}
        \put(22,0){\scriptsize{\cite{ho2020denoising_ddpm}}}
        \put(30,0){\scriptsize{\cite{song2020denoising_ddim}}}
        
        \put(37,0){\scriptsize{\cite{karras2020training_styleGan2ADA}}}
        \put(44,0){\scriptsize{\cite{karras2021alias_styleGan3}}}
        \put(51,0){\scriptsize{\cite{choi2020starganv2}}}
        \put(58,0){\scriptsize{\cite{li2021image_hisd}}}
        
        \put(66.5,0){\scriptsize{\cite{liu2019stgan}}}
        \put(73,0){\scriptsize{\cite{li2019faceshifter}}}
        
        \put(80.5,0){\scriptsize{\cite{liu2022pscc}}}
        \put(87,0){\scriptsize{\cite{liu2022pscc}}}
        \put(94.5,0){\scriptsize{\cite{wu2018busternet_copy_move_yue_wu}}}
        \end{overpic}
        \caption{}
        \label{fig_taxonomy_benchmark}
    \end{subfigure}
    \begin{subtable}[b]{0.20\linewidth}
        \centering
        \resizebox{1\textwidth}{!}{
            \small 
            \begin{tabular}{c|c}\hline
            Forgery Method&Image Source\\ \hline\hline
            DDPM~\cite{ho2020denoising_ddpm}& LSUN\\ \hline
            DDIM~\cite{song2020denoising_ddim}& LSUN\\ \hline
            GDM~\cite{nichol2021glide}& LSUN\\ \hline
            LDM~\cite{rombach2021highresolution_latent_diffusion}& LSUN\\ \hline
            StarGANv$2$~\cite{choi2020starganv2}& CelebaHQ\\ \hline
            HiSD~\cite{li2021image_hisd}& CelebaHQ\\ \hline
            StGANv$2$-ada~\cite{karras2020training_styleGan2ADA}& FFHQ, AFHQ\\ \hline
            StGAN$3$~\cite{karras2021alias_styleGan3}& FFHQ, AFHQ\\ \hline
            STGAN~\cite{liu2019stgan}& CelebaHQ\\ \hline
            Faceshifter~\cite{li2019faceshifter}& Youtube video\\ \hline
            \end{tabular}
        % \begin{tabular}{c|c|c|c|c}
        % \hline
        %  & \begin{tabular}[c]{@{}c@{}} CNN\\Syn.\end{tabular}
        %  & Editing & Mask & Year\\\hline
        %  \cite{mahfoudi2019defacto} & \textcolor{lightred}{\ding{56}} & \textcolor{citecolor}{\ding{52}} & \textcolor{citecolor}{\ding{52}} &$2019$\\ 
        %  \cite{stehouwer2019detection} & \textcolor{citecolor}{\ding{52}} & \textcolor{lightred}{\ding{56}} &
        %  \textcolor{citecolor}{\ding{52}} &$2019$\\ 
        % %  \cite{rossler2019faceforensics++v3} & \textcolor{citecolor}{\ding{52}} & \textcolor{lightred}{\ding{56}} &
        % %  \textcolor{lightred}{\ding{56}} & $2019$\\ \hline
        %  \cite{li_2020_CVPR} & \textcolor{citecolor}{\ding{52}} & \textcolor{lightred}{\ding{56}}& \textcolor{lightred}{\ding{56}}&$2020$\\ 
        %  \cite{Jiang_2020_CVPR} & \textcolor{citecolor}{\ding{52}} &
        %  \textcolor{lightred}{\ding{56}} &
        %  \textcolor{lightred}{\ding{56}} &$2020$\\ 
        %  \cite{IMD2020_wacv} & \textcolor{lightred}{\ding{56}} &
        %  \textcolor{citecolor}{\ding{52}} &
        %  \textcolor{citecolor}{\ding{52}} &$2020$\\ 
        %  \cite{he2021forgerynet} & 
        %  \textcolor{citecolor}{\ding{52}} & \textcolor{lightred}{\ding{56}} & \textcolor{citecolor}{\ding{52}} &$2021$\\
        %  \hline
        %  Ours & \textcolor{citecolor}{\ding{52}} & \textcolor{citecolor}{\ding{52}} & \textcolor{citecolor}{\ding{52}} &$2023$\\ \hline
        %  \end{tabular}
         }
        % \caption{The comparison to the previous proposed benchmark. We have added the newest diffusion models (\textit{i.e.} \cite{ho2020denoising_ddpm,song2020denoising_ddim,rombach2021highresolution_latent_diffusion,dhariwal2021diffusion_guided}).}
        \caption{}
        \label{fig_taxonomy_compare_table}
    \end{subtable}
    \begin{subfigure}[b]{0.22\linewidth}
    \centering
        \includegraphics[width=0.80\linewidth]{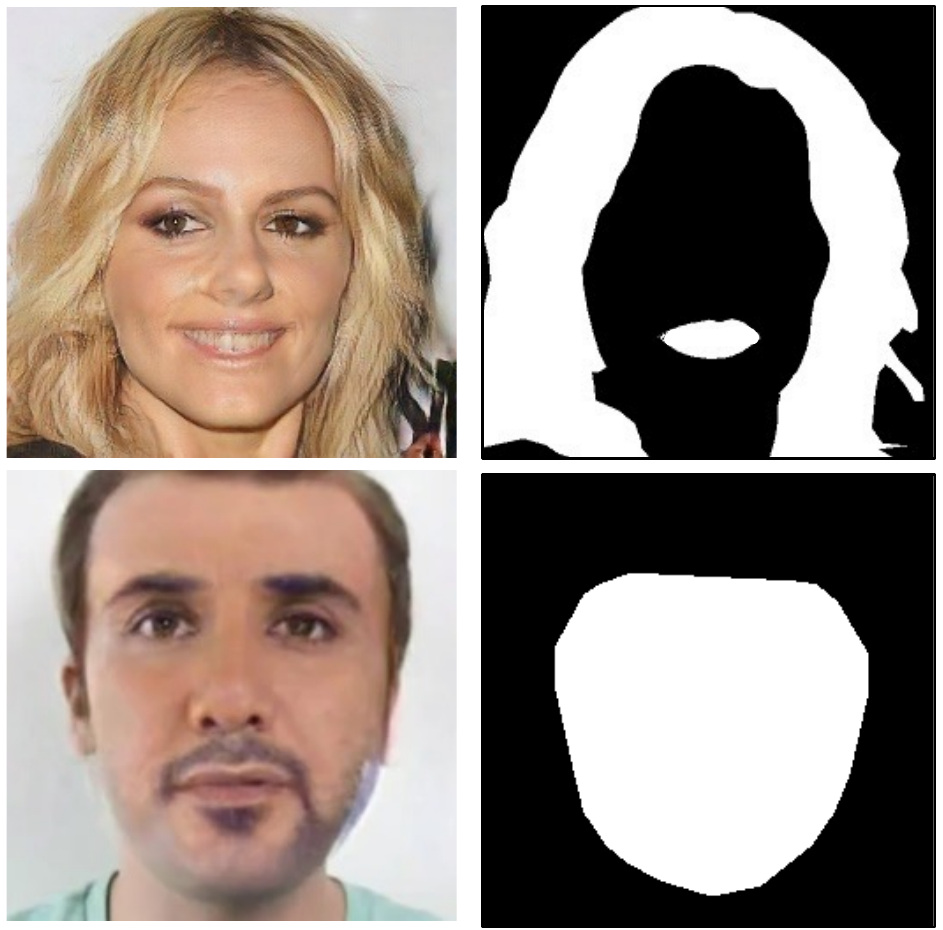}
        % \caption{We offer high resolution forgery masks on forged images.}
        \caption{}
        \label{fig_taxonomy_mask}
    \end{subfigure}
    \vspace{-2mm}
    \caption{Overview of HiFi-IFDL dataset. (a) At level $1$, we separate forged images into fully-synthesized and partial-manipulated. The second level discriminates different forgery methodologies, \textit{e.g.}, image editing, CNN-based partial manipulation, Diffusion or GANs. 
    Then, we separate images based on whether forgery methods are conditional or unconditional. 
    The final level refers to the specific forgery method. 
    (b) The table of forgery methods and images source that forgery methods are trained on. The dataset details can be found in the supplementary material. (c) We offer high resolution forgery masks on manipulated images.}
\end{figure*}
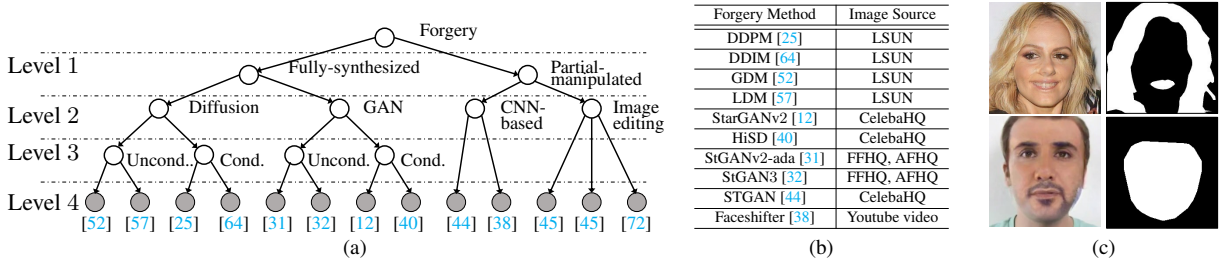
%\vspace{-5mm}
% \begin{figure}[t]
%     \centering
%     \begin{overpic}[width=\linewidth]{figure/manipulation_tree}
%     \put(-5,49){\footnotesize{Level 1}}
%     \put(-5,35){\footnotesize{Level 2}}
%     \put(-5,22){\footnotesize{Level 3}}
%     \put(-5,11){\footnotesize{Level 4}}
%     %%% END hierarchy
%     \put(25,58){\footnotesize{Manipulated Image}}
%     \put(20,40){\footnotesize{Editing}}
%     \put(58,40){\footnotesize{CNN-Manipulated}}
%     %%% END hierarchy
%     \put(29,27){\footnotesize{GAN}}  
%     \put(44.5,27){\footnotesize{DeepFakes}}
%     \put(71,27){\tiny{Diffusion}}
%     \put(30,-3){\tiny{~StarGANv2}~\cite{choi2020starganv2}}
%     \end{overpic}
%     \caption{The taxonomy of manipulation types. Improve the binary classification by predicting the entire taxonomy path towards the specific manipulated method. Meanwhile, this taxonomy turns manipulation attribute parsing in a closed-set classification problem with uncertainty.}
%     \label{fig:my_label}
% \end{figure}

%% file: section/03_01_benchmark.tex
\Section{Hierarchical Fine-grained IFDL dataset}\label{sec_benchmark}
We construct a fine-grained hierarchical benchmark, named HiFi-IFDL, to facilitate our study. 
HiFi-IFDL contains some most updated and representative forgery methods, for two reasons: 1) Image synthesis evolves into a more advanced era and artifacts become less prominent in the recent forgery method; 2) It is impossible to include all possible generative method categories, such as VAE~\cite{kingma2013auto_vae_gen} and face morphing~\cite{scherhag2019face_morphing}. 
So we only collect the most-studied forgery types (\textit{i.e.,} splicing) and the most recent generative methods (\textit{i.e.,} DDPM).

Specifically, HiFi-IFDL includes images generated from $13$ forgery methods spanning from CNN-based manipulations to image editing, as shown in the taxonomy of Fig.~\ref{fig_taxonomy_benchmark}. 
Each forgery method generates $100,000$ images. For the real images,  we select them from $6$ datasets (\textit{e.g.,} FFHQ~\cite{karras2019style}, AFHQ~\cite{choi2020starganv2}, CelebaHQ~\cite{lee2020maskgan_celebahq}, Youtube face~\cite{rossler2019faceforensics++v3}, MSCOCO~\cite{lin2014microsoft_coco}, and LSUN~\cite{yu2015lsun}). 
We either take the entire real image datasets or select $100,000$ images. 
Training, validation and test sets have $1,710$K, $15$K and $174$K images. 
While there are different ways to design a forgery hierarchy, our hierarchy starts at the root of an image being forged, and then each level is made more and more specific to arrive at the actual generator. 
Our work studies \emph{the impact of the hierarchical formulation to IFDL}. While different hierarchy definitions are possible, it is beyond the scope of this paper.

%% file: section/04_00_experiment.tex
\Section{Experiments}
We evaluate image forgery detection/localization (IFDL)  on $7$ datasets, and forgery attribute classification on HiFi-IFDL dataset. Our method is implemented on PyTorch and 
% trained with learning rate $\texttt{5e-5}$ on the feature extractor, and learning rate $\texttt{3e-5}$ on the localization module. 
% We train HiFi-Net 
trained with $400,000$ iterations, and batch size $16$ with $8$ real and $8$ forged images. The details can be found in the supplementary.
% We decrease learning rates by $0.95$ when the detection performance does not improve on validation split for $1,000$ iterations. 
\input{section/figure_table_latex/table_Tax_dataset_main_performance}
\input{section/figure_table_latex/table_img_editing_performance}

\SubSection{Image Forgery Detection and Localization}\label{ex_tax_IMDL}
\SubSubSection{HiFi-IFDL Dataset}\label{ex_data_IFDL}
Tab.~\ref{table_tax_IMDL} reports the different model performance on the HiFi-IFDL dataset, in which we use AUC and F$1$ score as metrics on both image-level forgery detection and pixel-level localization. Specifically, in Tab.~\ref{table_tax_det}, first we observe that the pre-trained CNN-detector~\cite{wang2020cnn} does not perform well because it is trained on GAN-generated images that are different from images manipulated by diffusion models. Such differences can be seen in Fig.~\ref{table_tax_artifacts}, where we visualize the frequency domain artifacts, by following the routine~\cite{wang2020cnn} that applies the high-pass filter on the image generated by different forgery methods. 
Similar visualization is adopted in~\cite{wang2020cnn,zhangxue2019detecting,corvi2022detection,ricker2022towards} also.
Then, we train both prior methods on HiFi-IFDL, and they again perform worse than our model: CNN-detector uses ordinary ResNet$50$, but our model is specifically designed for image forensics. 
Two-branch processes deepfakes video by LSTM that is less effective to detect forgery in image editing domain. Attention Xception~\cite{stehouwer2019detection} and PSCC~\cite{liu2022pscc} are proposed for facial image forgery and image editing domain, respectively. 
These two methods perform  worse than us by $9.3\%$ and $3.6\%$ AUC, respectively. We believe this is because our method can leverage localization results to help the image-level detection.

In Tab.~\ref{table_tax_loc}, we compare with previous methods which can perform the forgery localization. Specifically, the pre-trained OSN-detector~\cite{wu2022robust} and CatNet~\cite{kwon2022learning} do not work well on CNN-synthesized images in HiFi-IFDL dataset, since they merely train models on images manipulated by editing methods. 
Then, we use HiFi-IFDL dataset to train CatNet, but it still performs worse than ours: CatNet uses DCT stream to help localize area of splicing and copy-move, but HiFi-IFDL contains more forgery types (\textit{e.g.}, inpainting). Meanwhile, the accurate classification performance further helps the localization as statistics and patterns of forgery regions are related to different individual forgery method. For example, for the forgery localization, we achieve $2.6\%$ AUC and $2.0\%$ F$1$ improvement over PSCC. Additionally, the superior localization  demonstrates that our hierarchical fine-grained formulation learns more comprehensive forgery localization features than multi-level localization scheme proposed in PSCC. 

\SubSubSection{Image Editing Datasets}\label{ex_data_editing}
\vspace{-0.5mm}
Tab.~\ref{tab_img_editing} reports IFDL results for the image editing domain. We evaluate  on $5$ datasets: \textit{Columbia}~\cite{ng2009columbia}, \textit{Coverage}~\cite{wen2016coverage}, \textit{CASIA}~\cite{dong2013casia}, \textit{NIST$\textit{16}$}~\cite{NIST16} and \textit{IMD$20$}~\cite{IMD2020_wacv}. Following the previous experimental setup of~\cite{wu2019mantra,hu2020span,liu2022pscc,dong2022mvss,wang2022objectformer}, we pre-train the model on our proposed HiFi-IFDL and then fine-tune the pre-trained model on the \textit{NIST$\textit{16}$}, \textit{Coverage} and \textit{CASIA}. We also report the performance of HiFi-Net pre-trained on the same dataset as ~\cite{liu2022pscc}. 
% AUC and F$1$ are used for the detection and localization evaluation. 
Tab.~\ref{tab_img_editing_pre} reports the pre-trained model performance, in which our method achieves the best average performance. The ObjectFormer~\cite{wang2022objectformer} adopts the powerful transformer-based architecture and solely specializes in  forgery detection of the image editing domain, nevertheless its performance are on-par with ours.
%, even so it can only $0.3\%$ better AUC than ours.
In the fine-tune stage, our method achieves the best performance on average AUC and F$1$. Specifically, we only fall behind on  \textit{NIST16}, where AUC tends to saturate. 
We also report the image-level forgery detection results in Tab.~\ref{tab_img_editing_img_level}, achieving comparable results to ObjectFormer~\cite{wang2022objectformer}. We show qualitative results  in Fig.~\ref{fig_viz_manipulation}, where the manipulated region identified by our method can capture semantically meaningful object shape, such as the shapes of the tiger and squirrel. At last, we also offer the robustness evaluation in Tab.~\textcolor{red}{2} of the supplementary, showing our performance against various image transformations.
\input{section/figure_table_latex/table_DFFD_performance}
\SubSubSection{Diverse Fake Face Dataset}\label{ex_data_dffd}
% Then we examine the IFDL on the image editing domain, following the experiment setup defined in~\cite{wu2019mantra}, and compare to \cite{wu2019mantra,hu2020span,liu2022pscc,dong2022mvss,wang2022objectformer} on datasets including \textit{Columbia}~\cite{ng2009columbia}, \textit{Coverage}~\cite{wen2016coverage}, \textit{CASIA}~\cite{dong2013casia} and \textit{NIST$\textit{16}$}~\cite{NIST16}. After that, for the digital facial manipulated image, we compare to prior works~\cite{stehouwer2019detection,huang2020fakelocator} on DFFD dataset~\cite{stehouwer2019detection}. In the section~\ref{ex_mani_attribute}, 
% For the facial image forgery domain, 
We evaluate our method on the Diverse Fake Face Dataset (DFFD)~\cite{stehouwer2019detection}.
% which has different facial forgery methods, and real faces are from FFHQ~\cite{karras2019style} and CelebA~\cite{liu2015faceattributes_celeba}. We 
For a fair comparison, we follow the same experiment setup and metrics: IoU and pixel-wise binary classification accuracy (PBCA) for pixel-level localization, and AUC and PBCA for image-level detection. 
Tab.~\ref{tab_dffd} reports that our method obtains competitive performance on detection and the best localization performance on partial-manipulated images. More results are in the appendix.
% Considering that prior work~\cite{stehouwer2019detection} uses a powerful backbone architecture and is specific to the digital forgery domain, it further proves the generalization ability of our approach, achieving competitive performance even in the facial forgery domain.

%% file: section/figure_table_latex/table_Tax_dataset_main_performance.tex
\begin{table}[t]
\footnotesize
\begin{subtable}{1\linewidth}
    \centering
        \scalebox{1}{
        \begin{tabular}{c|cccc|cc}
        \hline
        \multirow{2}{*}{\begin{tabular}{c}
            \textbf{ Forgery}\\
            \textbf{ Detection}
        \end{tabular}}&\multicolumn{2}{c}{CNN-syn.}&\multicolumn{2}{c|}{Image Edit.}&\multicolumn{2}{c}{Overall}\\\cline{2-7}
        &AUC &F$1$ &AUC &F$1$ &AUC &F$1$\\ \hline
        CNN-det.$^*$~\cite{wang2020cnn}&$76.5$&$60.5$&$54.8$&$33.5$&$56.5$&$40.5$\\
        CNN-det.~\cite{wang2020cnn}&$92.3$&$90.0$&$87.0$&$74.7$&$90.1$&$83.7$\\
        Two-bran.~\cite{masi2020two}&$93.3$&$89.2$&$83.3$&$66.7$&$86.7$&$80.2$\\
        Att. Xce.~\cite{stehouwer2019detection}&$93.8$&$91.2$&$90.8$&$82.1$&$87.3$&$90.0$\\
        PSCC~\cite{liu2022pscc}&$94.6$&$93.2$&$90.7$&$82.3$&$93.2$&$91.3$\\
        \hline
        Ours&$\mathbf{97.0}$&$\mathbf{96.1}$&$\mathbf{91.5}$&$\mathbf{85.9}$&$\mathbf{96.8}$&$\mathbf{94.1}$\\ \hline
        \end{tabular}
        }
    \caption{CNN-detector~\cite{wang2020cnn} has $4$ variants with different augmentations, and we report the variant with the best performance. For Two-branch~\cite{masi2020two}, we implement this method with the help of its authors.\vspace{2mm}}
    \label{table_tax_det}
\end{subtable}

\begin{subtable}{1\linewidth}
    \centering
    \scalebox{1}{
    \begin{tabular}{c|cccc|cc}
    \hline
    \multirow{2}{*}{\begin{tabular}{c}
          \textbf{ Forgery}\\
        \textbf{ Localization}
    \end{tabular}}&\multicolumn{2}{c}{CNN-syn.}&\multicolumn{2}{c|}{Image Edit.}&\multicolumn{2}{c}{Overall}\\\cline{2-7}
    &AUC &F$1$ &AUC &F$1$ &AUC &F$1$\\ \hline
    OSN-det.$^*$~\cite{wu2022robust}&$51.4$&$38.8$&$83.2$&$70.1$&$79.4$&$56.5$\\    CatNet$^*$~\cite{kwon2022learning}&$48.6$&$31.9$&$86.1$&$79.4$&$78.3$&$65.1$\\
    CatNet~\cite{kwon2022learning}&$92.5$&$81.5$&$92.0$&$88.2$&$92.4$&$86.8$\\
    Att. Xce.~\cite{stehouwer2019detection}&$89.1$&$87.7$&$83.3$&$79.3$&$87.1$&$86.5$\\
    PSCC~\cite{liu2022pscc}&$94.3$&$96.8$&$91.1$&$86.5$&$92.7$&$94.9$\\
    \hline
    Ours&$\mathbf{98.4}$&$\mathbf{97.0}$&$\mathbf{93.0}$&$\mathbf{90.1}$&$\mathbf{95.3}$&$\mathbf{96.9}$\\ \hline
    \end{tabular}
    }
    \caption{OSN-det~\cite{wu2022robust} only releases pre-trained weights with the inference script, without the training script. \vspace{2mm}}
    \label{table_tax_loc}
\end{subtable}
\begin{subfigure}{1\linewidth}
    \centering
    \includegraphics[scale=0.3]{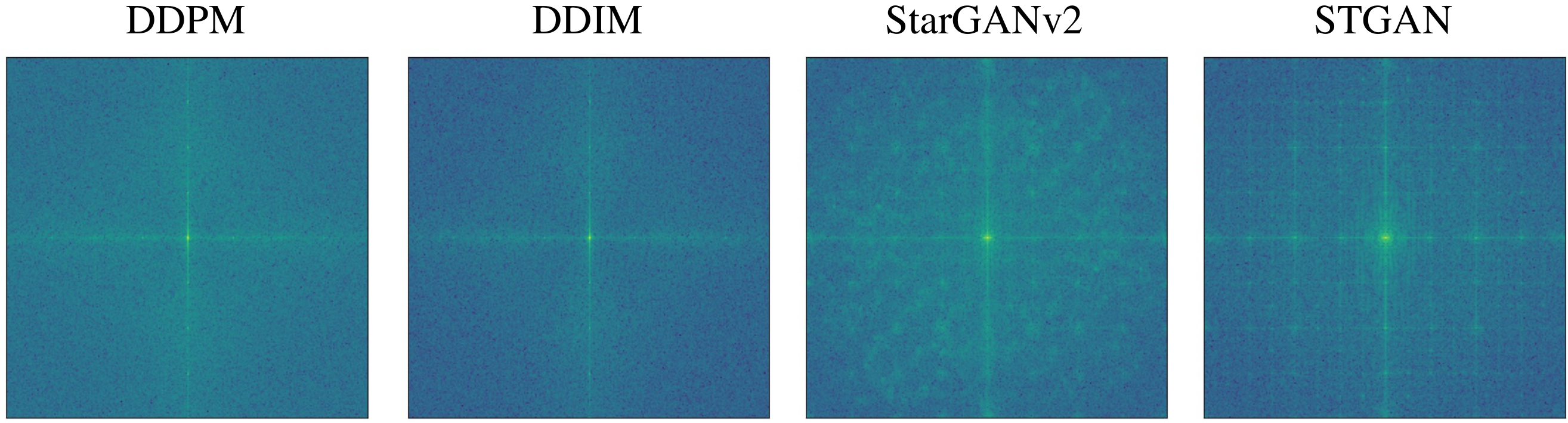}
    \caption{Frequency artifacts in different forgery methods. DDPM~\cite{ho2020denoising_ddpm} and DDIM~\cite{song2020denoising_ddim} do not exhbit the checkboard patterns~\cite{zhangxue2019detecting,wang2020cnn} observed in GAN-based methods, such as StarGAN-v$2$~\cite{choi2020starganv2} and STGAN~\cite{liu2019stgan}.}
    \label{table_tax_artifacts}
\end{subfigure}
\caption{IFDL Results on HiFi-IFDL. $*$ means we apply author-released pre-trained models. Models without $*$ mean they are trained on HiFi-IFDL training set. [\textbf{Bold}: best result].}
\label{table_tax_IMDL}
\end{table}
% \begin{figure}[h]
%     \centering
%     \includegraphics[scale=0.25]{results/artifacts.pdf}
%     \caption{Frequency artifacts in different forgery methods. DDPM~\cite{ho2020denoising_ddpm} and DDIM~\cite{song2020denoising_ddim} do not exhbit the checkboard patterns~\cite{zhang2019detecting,wang2020cnn} observed in GAN-based methods, such as StarGAN-v$2$~\cite{choi2020starganv2} and STGAN~\cite{liu2019stgan}.}
%     \label{table_tax_artifacts}
% \end{figure}

%%%%%%%%%%%%%%%%%%%%%%%%%%%%%%%%%%%%%%%%%%%%%%%%%%%%%%%%%%%%%%%%%%%%%%%
%%%%%%%%%%%%%%%%%OLD RESULT%%%%%%%%%%%%%%%%%%%%%%%%%%%%%%%%%%%%%%%%%%%%
% \begin{table}[t]
%     \footnotesize
%     \centering
%         \scalebox{1}{
%         \begin{tabular}{c|cc|cc}
%         \hline
%         \multirow{2}{*}{}&\multicolumn{2}{c|}{Detection}&\multicolumn{2}{c}{Localization}\\\cline{2-5}
%         &AUC &F$1$ &AUC &F$1$\\ \hline
%         Att. Xce.~\cite{stehouwer2019detection}&$87.3$&$90.0$&$87.1$&$86.5$\\
%         MVSNet~\cite{dong2022mvss}&$92.1$&$90.3$&$92.3$&$92.4$\\
%         PSCC~\cite{liu2022pscc}&$93.2$&$91.3$&$92.7$&$94.9$\\ \hline
%         Ours&$\mathbf{96.8}$&$\mathbf{94.1}$&$\mathbf{95.3}$&$\mathbf{96.9}$\\ \hline
%         \end{tabular}
%         }
%     \vspace{-2mm}
%     \caption{IFDL performance on HiFi-IFDL dataset. [Key: \textbf{Best}]\vspace{-3mm}.}
%     \vspace{-1mm}
%     \label{table_tax_IMDL}
% \end{table}

%% file: section/figure_table_latex/table_img_editing_performance.tex
\begin{figure*}[t]
    \centering
    \begin{overpic}[width=0.95\textwidth]{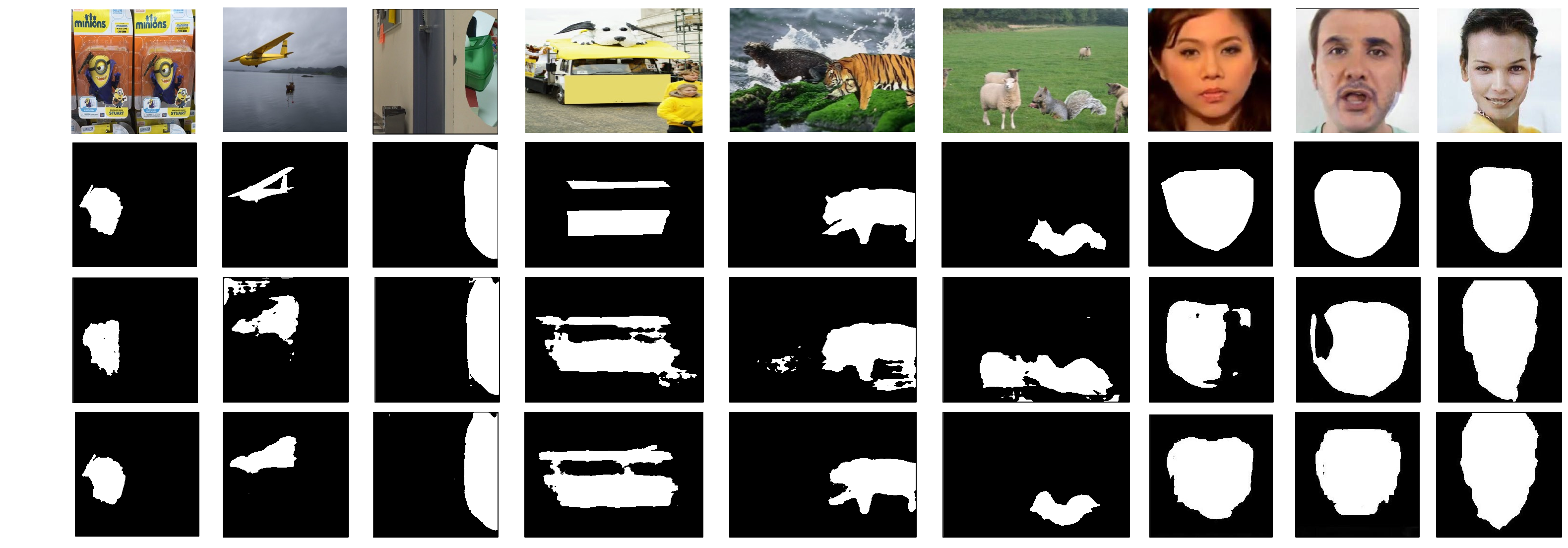}
    \put(-1,27.5){Image}
    \put(1,19){\small{GT}}
    \put(-0.5,12.5){\small{PSCC}}
    \put(0.5,10.5){\small{\cite{liu2022pscc}}}
    \put(0,4){\small{Ours}}
    \end{overpic}
    \vspace{-2mm}
    \caption{Qualitative results on different forged images. The first $6$ columns are from image editing methods whereas the last $3$ columns are images generated by Faceshifer~\cite{li2019faceshifter} and STGAN~\cite{liu2019stgan}.}
    \label{fig_viz_manipulation}
\end{figure*}
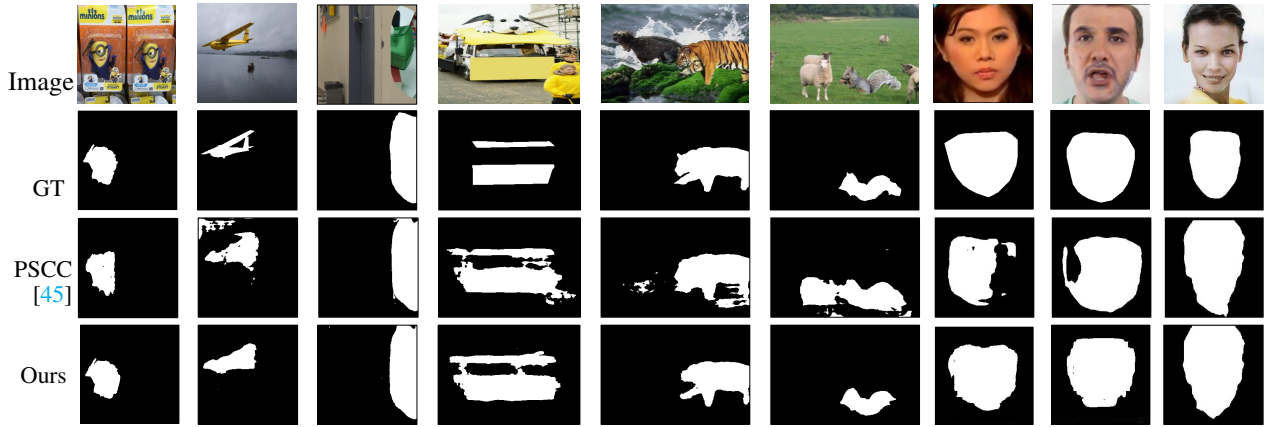
\begin{table*}[t]
\begin{subtable}{0.33\linewidth}
\centering
    \resizebox{0.95\textwidth}{!}{
        \begin{tabular}{c|ccccc|c}
        \hline
        \multirow{2}{*}{Localization}
        &Col.&Cov.&NI.$16$&CAS.&IM$20$&Avg.\\ \cline{2-7}
        &\multicolumn{6}{c}{\textit{Metric:} AUC($\%$) -- Pre-trained} \\ \hline
        ManT.\cite{wu2019mantra} &$82.4$&$81.9$&$79.5$&$81.7$ &$74.8$&$80.0$\\
        SPAN\cite{hu2020span}&$93.6$&$92.2$&$84.0$&$79.7$&$75.0$&$84.9$\\ 
        PSCC\cite{liu2022pscc}&$98.2$&$84.7$&$85.5$&$82.9$&$80.6$&$86.3$\\
        Ob.Fo.\cite{wang2022objectformer}&$95.5$&$\color{blue}92.8$&$\mathbf{\color{red}{87.2}}$&$84.3$&$82.1$&$88.3$\\
        \hline
        Ours$^*$ & $\color{blue}{98.3}$&$\mathbf{\color{red}{93.2}}$& $\color{blue}{87.0}$&$\color{blue}{85.8}$&$\color{blue}{82.9}$&$\color{blue}{89.4}$\\
        Ours & $\mathbf{\color{red}98.4}$&92.4& $86.9$&$\mathbf{\color{red}86.6}$&$\mathbf{\color{red}83.4}$&$\mathbf{\color{red}89.6}$\\  \hline
        \end{tabular}
    }
    \caption{}
\label{tab_img_editing_pre}
\end{subtable}\hfill
\begin{subtable}{0.42\linewidth}
\centering
    \resizebox{1\textwidth}{!}{
        \begin{tabular}{c| c c c | c}
        \hline
        \multirow{2}{*}{Localization}
        &Cov.&CAS.&NI.$16$&Avg.\\ \cline{2-5}
        &\multicolumn{4}{c}{\textit{Metric:} AUC($\%$) / F$1$($\%$) -- Fine-tuned} \\ \hline
        SPAN\cite{hu2020span}&$93.7/55.8$&$83.8/40.8$&$96.1/58.2$&$91.2/51.6$\\
        PSCC\cite{liu2022pscc}&$94.1/72.3$&$87.5/55.4$&$\color{red}\mathbf{99.6}$/$81.9$&$93.7/69.8$\\
        Ob.Fo.\cite{wang2022objectformer}&$\color{blue}95.7$/$\color{blue}75.8$&$\color{blue}88.2$/$\color{blue}57.9$&$\color{blue}99.6$/$\color{blue}82.4$&$\color{blue}{94.5}$/$\color{blue}72.0$\\
        \hline
        % Ours$^*$ & $\mathbf{\color{red}{96.2}}$/$\color{blue}{79.3}$&$\color{blue}{88.2}$/$\color{blue}60.6$ & $98.3$/$\color{blue}84.8$&$94.2$/$\color{blue}74.9$\\  
        Ours & $\color{red}{\mathbf{96.1}}$/$\mathbf{\color{red}{80.1}}$&$\color{red}\mathbf{88.5}$/$\color{red}\mathbf{61.6}$ & $98.9$/$\color{red}\mathbf{85.0}$&$\color{red}\mathbf{94.6}$/$\color{red}\mathbf{75.5}$\\
        \hline
        \end{tabular}
    }
    \caption{}
\label{tab_img_editing_ft}
\end{subtable}\hfill
\begin{subtable}{0.21\linewidth}
\centering
    \resizebox{1.07\textwidth}{!}{
        \begin{tabular}{c| c c}
        \hline
        Detection&AUC(\%)&F$1$(\%)\\ \hline
        ManT.\cite{wu2019mantra}&$59.9$&$56.7$\\
        SPAN\cite{hu2020span}&$67.3$&$63.8$\\
        PSCC\cite{liu2022pscc}&$99.5$&$97.1$\\
        Ob.Fo.\cite{wang2022objectformer}&$\color{red}\mathbf{99.7}$&$\color{blue}97.3$\\
        \hline
        % Ours$^*$ &$\color{blue}99.6$&$\color{red}\mathbf{97.7}$\\
        Ours & $\color{blue}99.5$&$\color{red}\mathbf{97.4}$\\  
        \hline
        \end{tabular}
    }
    \caption{}
\label{tab_img_editing_img_level}
\end{subtable}\hfill
% \vspace{1mm}
% \vspace{-3mm}
\caption{ IFDL results on the image editing. (a) Localization performance of the pre-train model. (b) Localization performance of the fine-tuned model. (c)  Detection performance on \textit{CASIA} dataset. All results of prior works are ported from~\cite{wang2022objectformer}. [Key: \textcolor{red}{\textbf{Best}}; \textcolor{blue}{Second Best}; \textit{Ours$^*$} uses the same pre-trained dataset as~\cite{liu2022pscc}, and \textit{ours} is pre-trained on HiFi-IFDL].}
\label{tab_img_editing}
% \vspace{-2mm}
\end{table*}
% \vspace{-3mm}

%% file: section/figure_table_latex/table_DFFD_performance.tex
\begin{table}[t]
\footnotesize
\begin{subtable}{0.45\linewidth}
    \centering
    \resizebox{1\textwidth}{!}{
        \begin{tabular}{c| c}
        \hline
        \multicolumn{2}{c}{\textit{Metric:} IoU / PBCA} \\
        \hline
        Att.Xce.~\cite{stehouwer2019detection} &$0.401/0.786$\\
        \hline
        Ours & $\mathbf{0.411}/\mathbf{0.801}$ \\  \hline
        % \multicolumn{2}{c}{\textit{Metric:} IINC ($\downarrow$) (real / fake / par.)} \\
        % \hline
        % \cite{stehouwer2019detection} & $0.011/0.090/0.373$ \\
        % \hline
        % Ours & $\textbf{0.010}/0.050/\textbf{0.273}$\\\hline
        \end{tabular}
    }
    \caption{localization}
\end{subtable}
\begin{subtable}{0.45\linewidth}
    \centering
    \resizebox{1\textwidth}{!}{
        \begin{tabular}{c|c}
        \hline
        \multicolumn{2}{c}{\textit{Metric:} AUC/PBCA} \\
        \hline
        Att.Xce.~\cite{stehouwer2019detection} &$\mathbf{99.69}/88.44$\\
        \hline
        Ours & $99.45/\mathbf{88.50}$ \\  \hline
        \end{tabular}
    }
    \caption{detection}
\end{subtable}
\vspace{-2mm}
\caption{IFDL results on DFFD dataset. [Key: \textbf{Best}]}
\vspace{-3mm}
\label{tab_dffd}
\end{table}

%% file: section/04_01_experiment_ablation.tex
\subsection{Ablation Study}\label{ex_ablation}
% We report the ablation study in Tab.~\ref{table_ablation}. 
In row~$1$ and $2$ of Tab.~\ref{table_ablation}, we first ablate the $\mathcal{L}_{loc}$ and $\mathcal{L}_{cls}$, removing which causes  large performance drops on the detection ($24.1\%$ F$1$) and localization ($29.3\%$ AUC), respectively. Also, removing $\mathcal{L}_{cls}$ harms localization by $1.9\%$ AUC and F$1$. 
This shows that fine-grained classification improves the localization, as the fine-grained classification features serve as a prior for localization. 
% Moreover, we visualize the t-SNE low-dimensional space generated by $\mathcal{L}_{loc}$ in Fig.~\ref{fig_confusion_matrix}\textcolor{red}{(a)}. 
We evaluate the effectiveness of performing fine-grained classification at different hierarchical levels. 
In the $3$th row, we only keep the $4$th level fine-grained classification in the training, which causes a sensible drop of  performance in detection ($3.7\%$ AUC) and localization ($2.8\%$ AUC). 
In the $4$th row, we perform the fine-grained classification % separately on different levels but 
without forcing the dependency between layers of~Eq.~\ref{eq_conditional_prob}. 
This % modification 
impairs the learning of hierarchical forgery attributes and causes a drop of $3.6\%$ AUC in the detection.
% Such phenomena show that our fine-grained hierarchical formulation indeed helps model learn more comprehensive feature towards the IFDL. 
Lastly, we ablate the PConv in the $5$th row, making model less effective for detection.
\input{section/figure_table_latex/table_Tax_dataset_ablation}
\input{section/figure_table_latex/figure_res_mani_attribute}
\subsection{Forgery Attribute performance}\label{ex_mani_attribute}

\begin{figure}[t]
    \centering
    \begin{overpic}[width=0.44\textwidth]{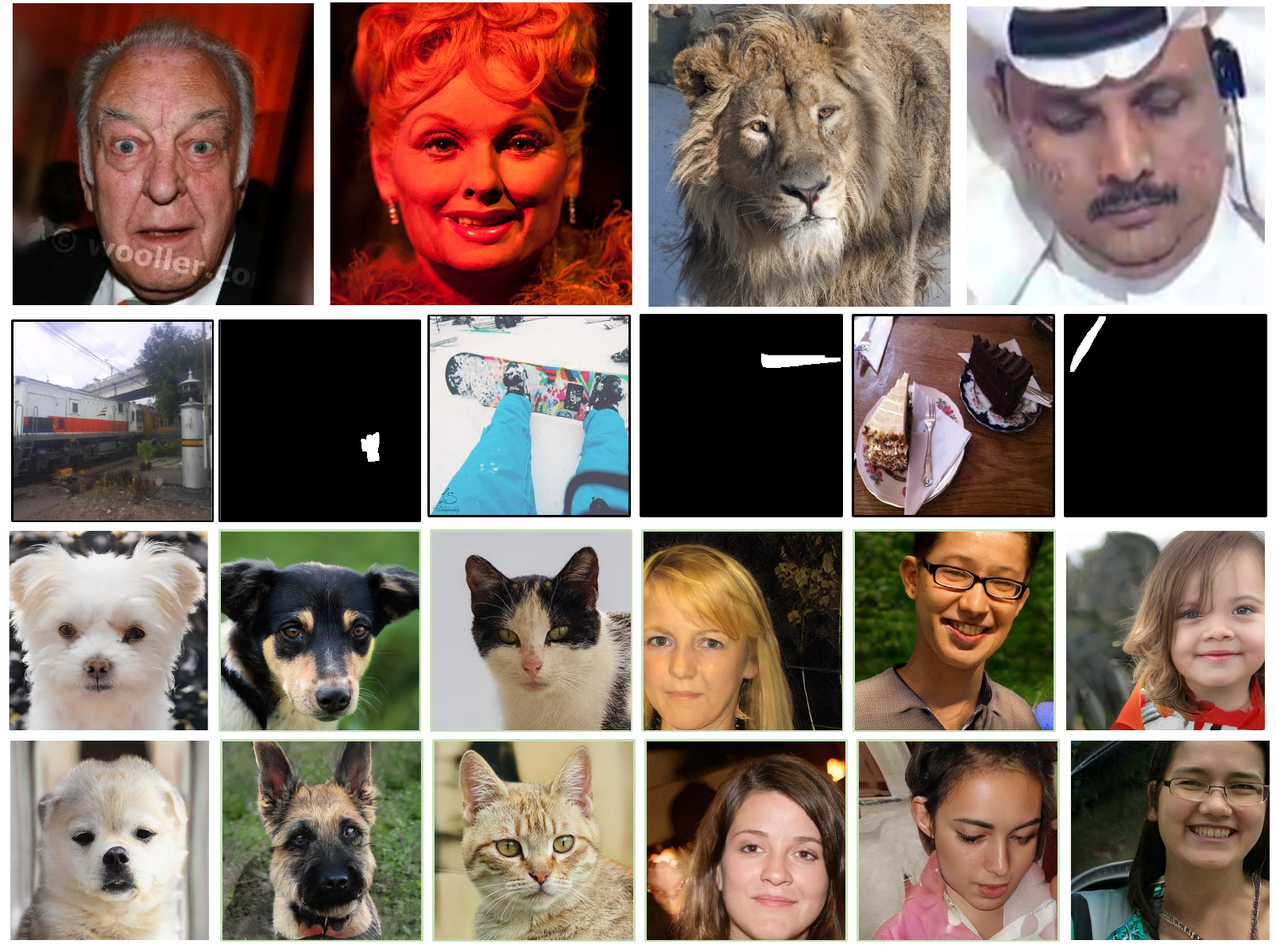}
    \put(-5,60){(a)}
    \put(-5,40){(b)}
    \put(-5,15){(c)}
    \end{overpic}
    \caption{Three failure scenarios: (a) real images. (b) inpainting images with small removal regions. (c) images generated from different styleGANv$2$ada~\cite{karras2020training_styleGan2ADA} and styleGANv$3$~\cite{karras2021alias_styleGan3}, as shown in the last two rows, respectively.}
    \label{fig_gallery}
\end{figure}

We perform the fine-grained classification among real images and $13$ forgery categories on $4$ different levels, and the most challenging scenario is the fine-grained classification on the $4$~th level. % reported in Tab.~\ref{tab_fg_performance}.
The result is reported in Tab.~\ref{tab_fg_performance}. Specifically, we train HiFi-Net $4$ times, and at each time only classifies the fine-grained forgery attributes at one level, denoted as \textit{Baseline}. 
Then, we train a HiFi-Net to classify all $4$ levels but without the hierarchical dependency via Eq.~\ref{eq_conditional_prob}, denoted as \textit{multi-scale}. Also, we compare to the pre-trained image attribution works~\cite{asnani2021reverse,yu2019attributing_image_attribute}. 
Also, it has been observed in Fig.~\ref{fig_confusion_matrix} that we fail on $3$ scenarios
1) Some real images have watermarks, extreme lightings, and distortion. 
2) Inpainted images have small forgery regions.
3) styleGANv$2$-ada~\cite{karras2020training_styleGan2ADA} and styleGAN$3$~\cite{karras2021alias_styleGan3} can produce highly similar images.
Fig.~\ref{fig_gallery} shows failure cases.

%% file: section/figure_table_latex/table_Tax_dataset_ablation.tex
\begin{table}[t]
\footnotesize
\centering
    \scalebox{1.0}{
    \begin{tabular}{ccc|cc|cc}
    \hline
    % \multirow{2}{*}{Method}
    
    % &&&\multicolumn{2}{c|}{Detection}&\multicolumn{2}{c}{Localization}\\ \cline{4-7}
    % &Mod.&Loss&AUC &F$1$ &AUC &F$1$\\ \hline\hline
    
    \multirow{2}{*}{}&\multirow{2}{*}{Method}&\multirow{2}{*}{Loss}
    &\multicolumn{2}{c|}{Detection}&\multicolumn{2}{c}{Localization}\\ \cline{4-7}
    &&&AUC &F$1$ &AUC &F$1$\\ \hline\hline
    
    \textit{Full}&\textbf{M},\textbf{L},\textbf{P}&$\mathcal{L}_{cls}$, $\mathcal{L}_{loc}$&$\mathbf{96.8}$&$\mathbf{94.1}$&$\mathbf{95.3}$&$\mathbf{96.9}$\\ \hline
    \textit{1}&\textbf{M},\textbf{L},\textbf{P}&$\mathcal{L}_{loc}$&$65.0$&$70.0$&$93.4$&$95.0$\\ \hline
    \textit{2}&\textbf{M},\textbf{L},\textbf{P}&$\mathcal{L}_{cls}$&$95.8$&$92.4$&$66.0$&$58.0$\\ \hline
    \textit{3}&\textbf{M},\textbf{L},\textbf{P}&$\mathcal{L}^{4}_{cls}$,$\mathcal{L}_{loc}$&$93.1$&$91.7$&$92.5$&$93.9$\\ \hline
    \textit{4}&\textbf{M},\textbf{L},\textbf{P}&$\mathcal{L}^{ind}_{cls}$,$\mathcal{L}_{loc}$&$93.2$&$92.8$&$93.2$&$94.8$\\ \hline
    \textit{5}&\textbf{M},\textbf{L}&$\mathcal{L}_{cls},\mathcal{L}_{loc}$&$96.6$&$93.0$&$94.8$&$96.0$\\ \hline
    \end{tabular}
    }
\caption{Ablation study. \textbf{M}, \textbf{L}, and \textbf{P} represent the multi-branch classification module, localization module and Pconv operation, respectively. $\mathcal{L}_{cls}$ and $\mathcal{L}_{loc}$ are classification and localization loss, respectively. $\mathcal{L}^{4}_{cls}$ and $\mathcal{L}^{ind}_{cls}$ denote we only perform the fine-grained classification on $4^{th}$ level and classification without hierarchical path prediction. [Key: \textbf{Best}]}
\label{table_ablation}
\end{table}

%% file: section/figure_table_latex/figure_res_mani_attribute.tex
\begin{figure}[t]
    \centering
    \begin{subtable}[b]{0.5\linewidth}
        \centering
        \small
        \begin{tabular}{c|c}\hline
             &F$1$ score\\ \hline\hline
        Attr.~\cite{yu2019attributing_image_attribute}&$31.11$\\\hline          FEN~\cite{asnani2021reverse}&$48.00$\\\hline
        Baseline&$84.62$\\\hline
        Multi-scale&$85.37$\\ \hline
        HiFi-Net&$\mathbf{87.63}$\\\hline
        \end{tabular}        
        \caption{}
        \label{tab_fg_performance}
    \end{subtable} \hfill
    \begin{subfigure}[b]{0.46\linewidth}
        \centering
        \resizebox{1\textwidth}{!}{
            \begin{overpic}[width=1\linewidth]{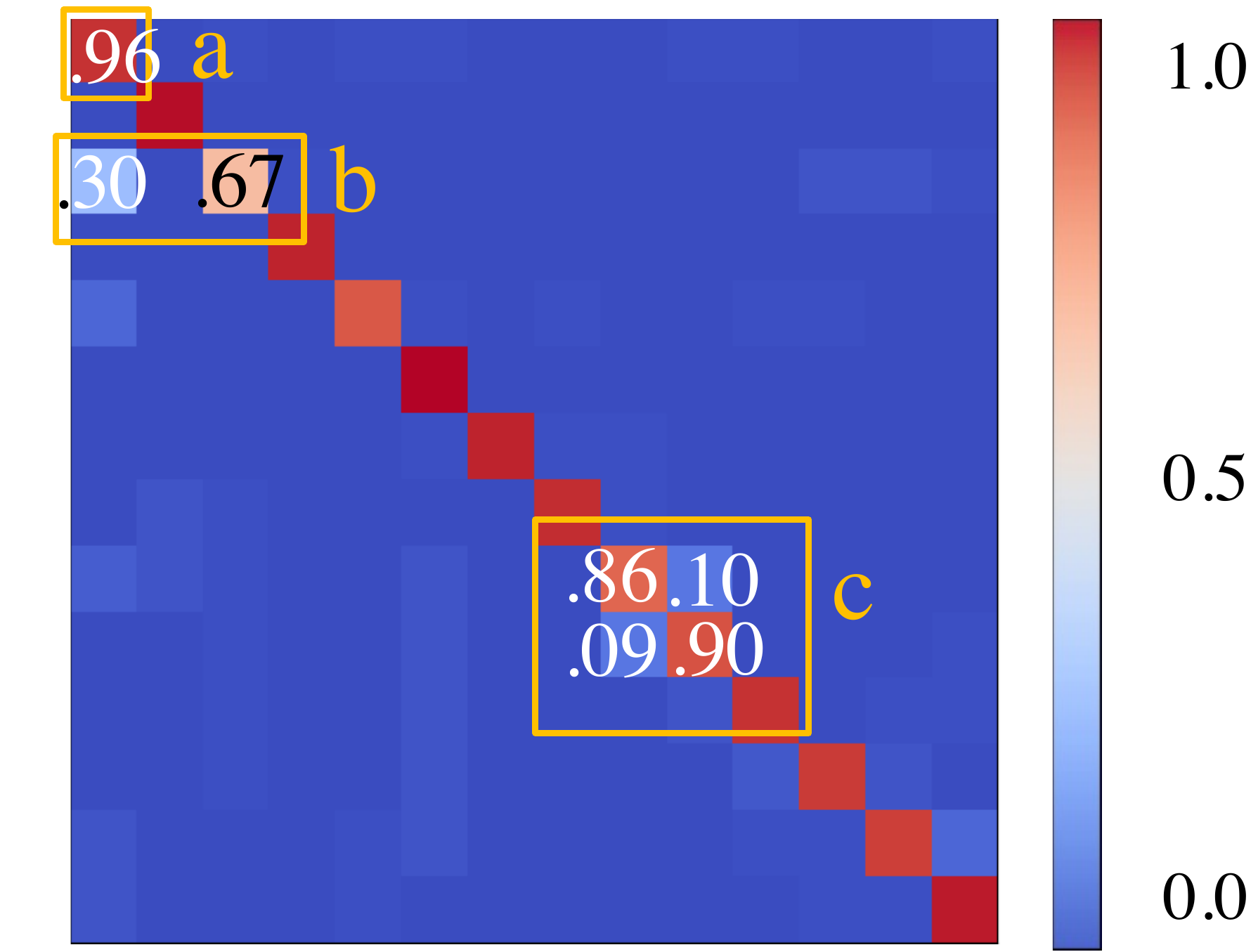}
                \put(-13,14){\small\rotatebox{90}{True Label}}
            \put(-10,72){\scriptsize{Real}}
            \put(-13,65.5){\scriptsize{Splice}}
            \put(-15,59.5){\scriptsize{Inpaint}}
            \put(-6,55){\tiny{\cite{liu2019stgan}}}
            \put(-6,49){\tiny{\cite{liu2019stgan}}}
            \put(-6,44){\tiny{\cite{li2019faceshifter}}}
            \put(-6,38){\tiny{\cite{choi2020starganv2}}}
            \put(-6,33){\tiny{\cite{li2021image_hisd}}}
            \put(-6,28){\tiny{\cite{karras2020training_styleGan2ADA}}}
            \put(-6,23){\tiny{\cite{karras2021alias_styleGan3}}}
            \put(-6,18){\tiny{\cite{ho2020denoising_ddpm}}}
            \put(-6,13){\tiny{\cite{song2020denoising_ddim}}}
            \put(-6,8){\tiny{\cite{nichol2021glide}}}
            \put(-6,3){\tiny{\cite{rombach2021highresolution_latent_diffusion}}}
            \end{overpic}
        }
        \caption{}
        \label{fig_confusion_matrix}
    \end{subfigure}
    \caption{(a) The forgery attribute classification results. The improvement over previous works~\cite{yu2019attributing_image_attribute,asnani2021reverse} is because the previous works only learn to attribute CNN-synthesized images, yet do not consider attributing image editing methods. (b) The confusion matrix of  forgery attribute classification at level $4$, where a, b and c represent three scenarios of classification failures. The numerical value indicates the accuracy. See  Sec.~\ref{ex_mani_attribute} and Fig.~\ref{fig_gallery} for details.}
\end{figure}

%% file: section/05_conclusion.tex
\Section{Conclusion}
% Being motivated by conceiving a more well-rounded IFDL algorithm, 
In this work, we develop a method for both CNN-synthesized and image editing forgery domains.
% Our method is distinct to previous IFDL work which treats IFDL in image editing or CNN-manipulated domains as two separate topics. 
We formulate the IFDL as a hierarchical fine-grained classification problem which requires the algorithm to classify the individual forgery method of given images, via predicting the entire hierarchical path. Also, HiFi-IFDL dataset is proposed to further help the community in developing forgery detection algorithms.
% We also propose and release the HiFi-IFDL dataset which can further help the community in developing forgery detection algorithms.
% a unified solution for detecting forged images.
%Specifically, this formulation encourages the algorithm to capture more detailed distinction between different forgery methods, thereby improve the overall IFDL performance. We believe that researches always benefit in exploring the subtle difference between intra-class categories, %not only in the forgery detection domain, also other detection topic. 
% The second usage of our hierarchical fine-grained formulation is, we convert the forgery attribute into a close-form prediction problem, while keeping the hierarchical dependency. This helps the algorithm becomes versatile, in that it is optimized towards the forgery detection but also achieves the purpose of forgery attribute classification.
% \vspace{-2mm}
\paragraph{Limitation} 
% As shown in the Sec.~\ref{ex_generalization}, 
% Although our method performs well on the conventional image editing manipulation, it actually generalizes poorly on the image partially manipulated by diffusion-based methods. 
Please refer to the supplementary Sec.~\textcolor{red}{2}: the model that performs well on the conventional image editing can generalize poorly on diffusion-based inpainting method. 
Secondly, we think it is possible improve the IFDL learning via the larger forgery dataset.
% more forgery methods.
% Our plan is to collect over $1,000$ different forgery methods in the future, to further improve the IFDL study.
% Although we achieve the state-of-the-art performance on the IFDL task, it is actually possible to explore different, or trainable taxonomy or tree-structure that helps neural network capture the forgery attribute. 

%% file: section/10_appendix.tex
\subsection{Dataset Collection Details}
\begin{figure*}[ht]
    \centering
    \begin{overpic}[width=\textwidth]{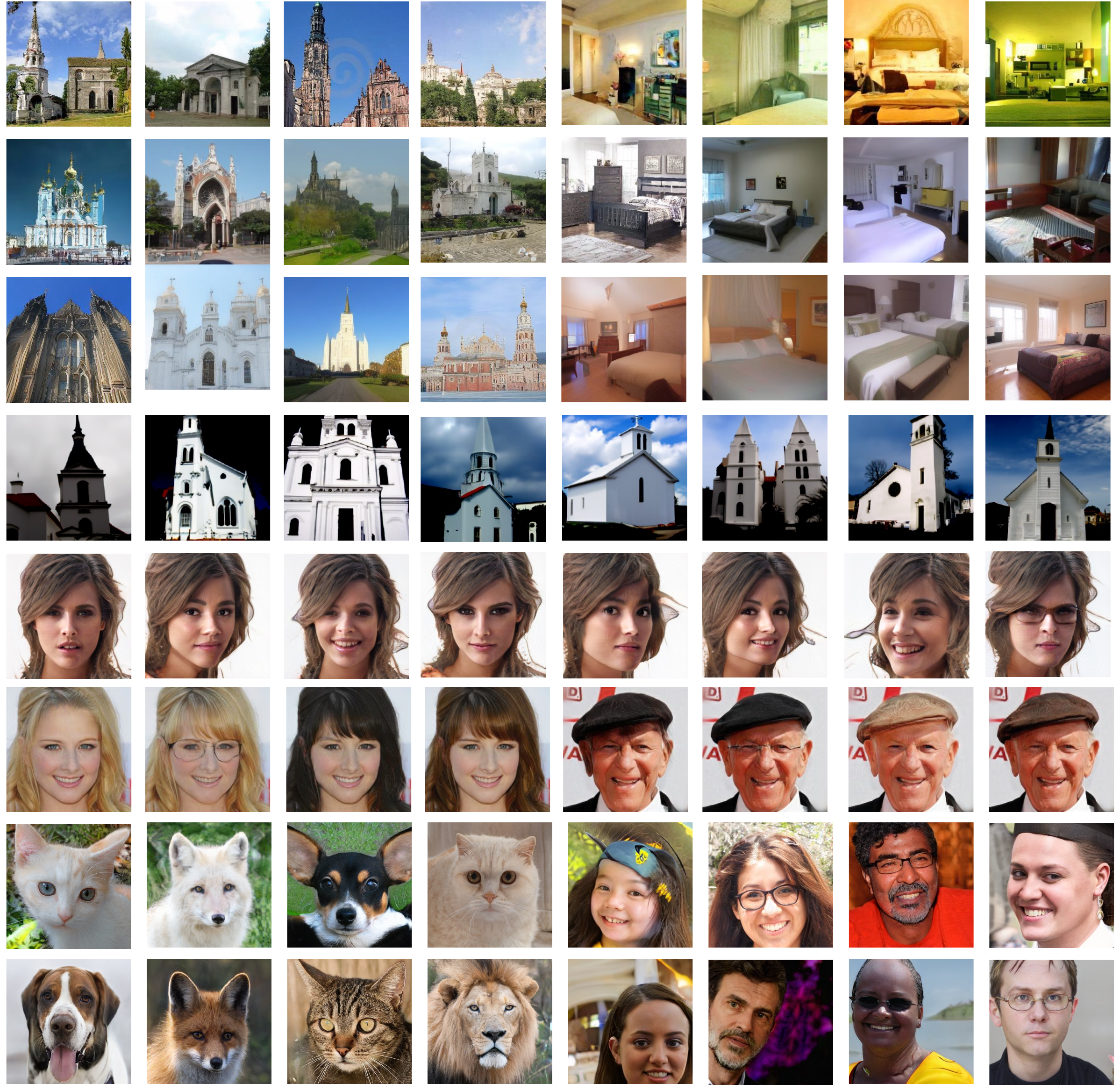}
    \end{overpic}
    \vspace{-2mm}
    \caption{The samples from the proposed HiFi-IFDL dataset. From top to bottom, the images are generated by DDPM, DDIM, LDM, GDM, StarGANv$2$, HiSD, StyleGANv$2$ada, StyleGANv$3$.}
    \label{fig_supp_gallery_dataset}
\end{figure*}

\begin{figure*}[t]
    \centering
    \begin{overpic}[width=1.\textwidth]{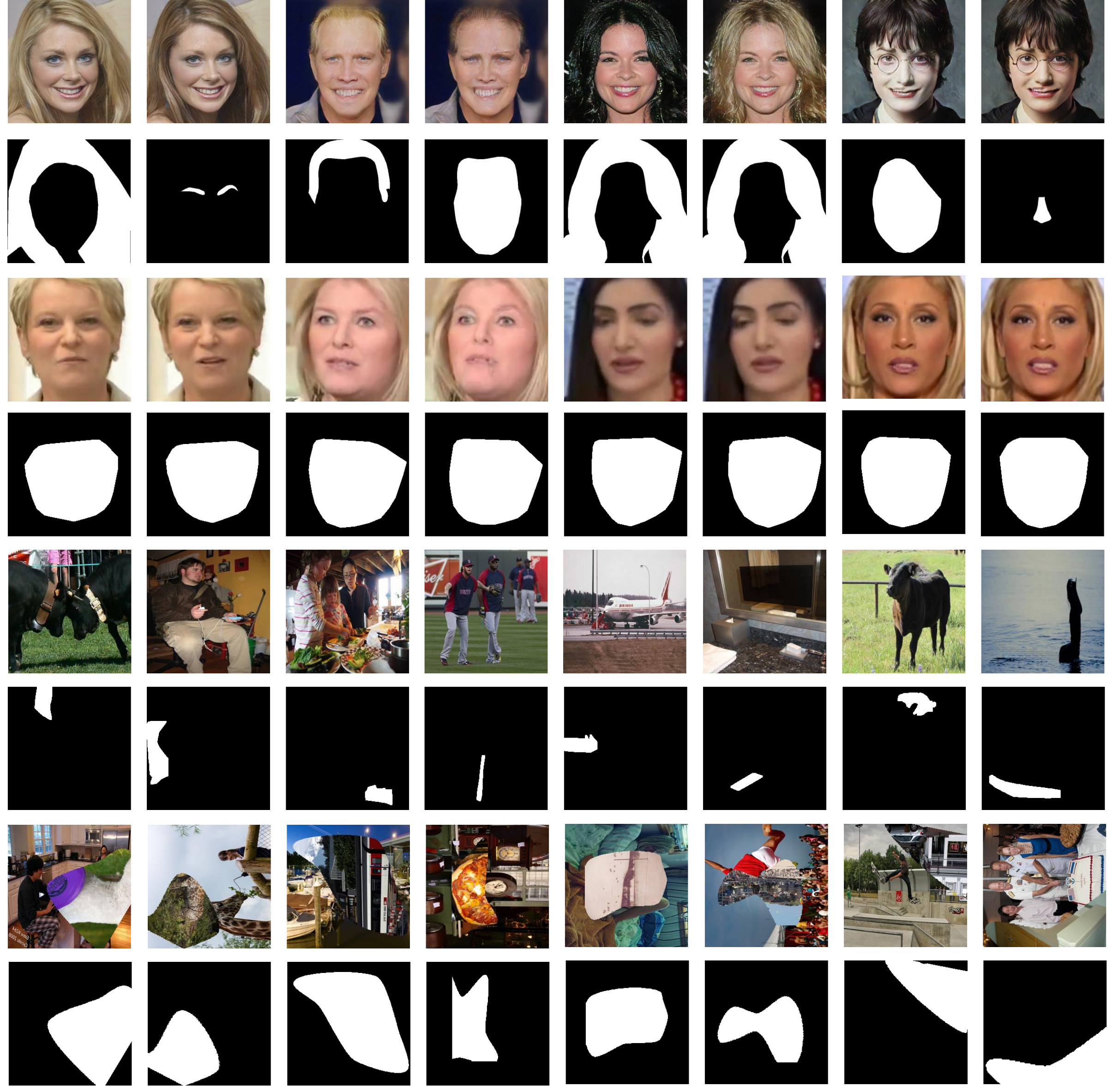}
    \end{overpic}
    \vspace{-2mm}
    \caption{Additional samples from the proposed HiFi-IFDL dataset. From top to bottom, the images are generated by STGAN, Faceshifter, and two image editing methods.}
    \label{fig_supp_gallery_v2}
\end{figure*}

Table~\ref{tab_dataset} reports all the forgery methods used in our dataset. 
In the last column, the table shows if the method used to generate the manipulated images is pre-trained, self-trained, or we used the released images. 
In Fig.~\ref{fig_supp_gallery_dataset} and Fig.~\ref{fig_supp_gallery_v2}, we show several examples taken from our dataset that represents a variety of objects, scenes, faces, animals. 
The real image dataset is the combination of LSUN~\cite{yu2015lsun}, CelebaHQ~\cite{lee2020maskgan_celebahq}, FFHQ~\cite{karras2019style}, AFHQ~\cite{choi2020starganv2}, MSCOCO~\cite{lin2014microsoft_coco} and real face images in face forensics~\cite{rossler2019faceforensics++v3}. We either take the entire dataset or randomly select $100$k images from these real datasets.

\begin{table}[ht]
\small
\centering
\begin{tabular}{c|c|c|c}\hline
Forgery Method&Image Source&Images $\#$&Source\\ \hline
DDPM~\cite{ho2020denoising_ddpm}& LSUN& $100$k & pre-trained\\ \hline
DDIM~\cite{song2020denoising_ddim}& LSUN& $100$k & pre-trained\\ \hline
GDM.~\cite{nichol2021glide}& LSUN& $100$k & pre-trained\\ \hline
LDM.~\cite{rombach2021highresolution_latent_diffusion}& LSUN& $100$k & pre-trained\\ \hline
StarGANv$2$~\cite{choi2020starganv2}& CelebaHQ& $100$k & pre-trained\\ \hline
HiSD~\cite{li2021image_hisd}& CelebaHQ& $100$k & pre-trained\\ \hline
StGANv$2$-ada~\cite{karras2020training_styleGan2ADA}& FFHQ, AFHQ& $100$k & pre-trained\\ \hline
StGAN$3$~\cite{karras2021alias_styleGan3}& FFHQ, AFHQ& $100$k & pre-trained\\ \hline
STGAN~\cite{liu2019stgan}& CelebaHQ& $100$k & self-train\\ \hline
Faceshifter~\cite{li2019faceshifter}& Youtube video& $100$k & released\\ \hline
\hline
\end{tabular}
\vspace{-2mm}
\caption{The details of the collected dataset. Each column in order shows forgery method; the image source used for the generation; the image number; if the images are generated with a pre-trained/self-trained models or released images.}
\label{tab_dataset}
\end{table}

\subsection{Generalization Performance}
\begin{figure}[t]
    \centering
    \vspace{-3mm}
    \begin{overpic}[scale=0.35]{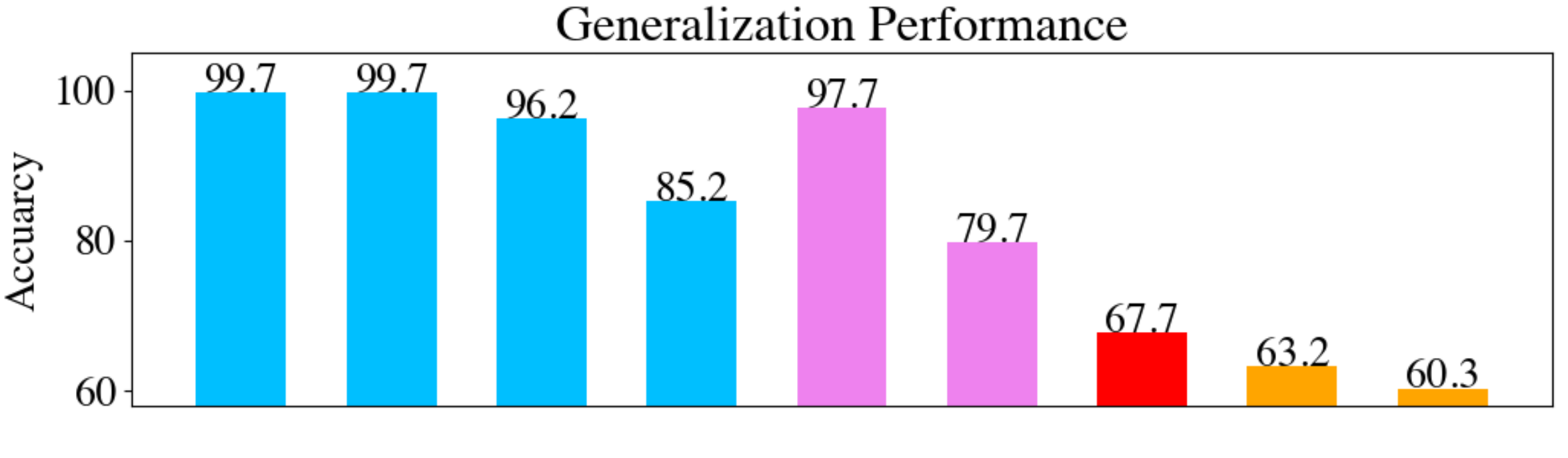}
    \put(10,0){\tiny{ProGAN}}
    \put(12,-3){\scriptsize{\cite{karras2018progressive}}}
    \put(20,0){\tiny{StyGAN}}
    \put(21,-3){\scriptsize{\cite{karras2019style}}}
    \put(30,0){\tiny{DPGAN}}
    \put(31,-3){\scriptsize{\cite{wang2022diffusion}}}
    \put(40,0){\tiny{DSGAN}}
    \put(41,-3){\scriptsize{\cite{wang2022diffusion}}}
    \put(50,0){\tiny{ADM}}
    \put(51,-3){\scriptsize{\cite{dhariwal2021diffusion}}}
    \put(60,0){\tiny{PNDM}}
    \put(61,-3){\scriptsize{\cite{liu2022pseudo}}}
    \put(70,0){\tiny{SiWM-v2}}
    \put(71,-3){\scriptsize{\cite{guo2022multi}}}
    \put(80,0){\tiny{RePaint}}
    \put(80,-3){\scriptsize{\cite{lugmayr2022repaint}}}
    \put(90,0){\tiny{LDM}}
    \put(90,-3){\scriptsize{\cite{rombach2021highresolution_latent_diffusion}}}
    \end{overpic}
    \vspace{-3mm}
    \caption{The image-level forgery detection accuracy on images generated by unseen GAN~(\textcolor{cyan}{\rule{0.4cm}{0.25cm}}) and diffusion~(\textcolor{pinkpearl}{\rule{0.4cm}{0.25cm}}) methods, and unseen domain real images~(\textcolor{red}{\rule{0.4cm}{0.25cm}}). The pixel-level localization accuracy on images inpainted by unseen diffusion model~(\textcolor{princetonorange}{\rule{0.4cm}{0.25cm}}). From left to right, first $6$ methods produce \emph{bedroom} in LSUN~\cite{yu2015lsun}. The SiW-Mv$2$ contains real human face. Images generated from the last two diffusion model inpainting methods, are human face and general objects. All these images are either obtained directly from the open source github or the pre-trained weight.}
    \label{fig_unseen_gen}
\end{figure}
\begin{figure}[t]
    \centering
    \includegraphics[scale=0.58]{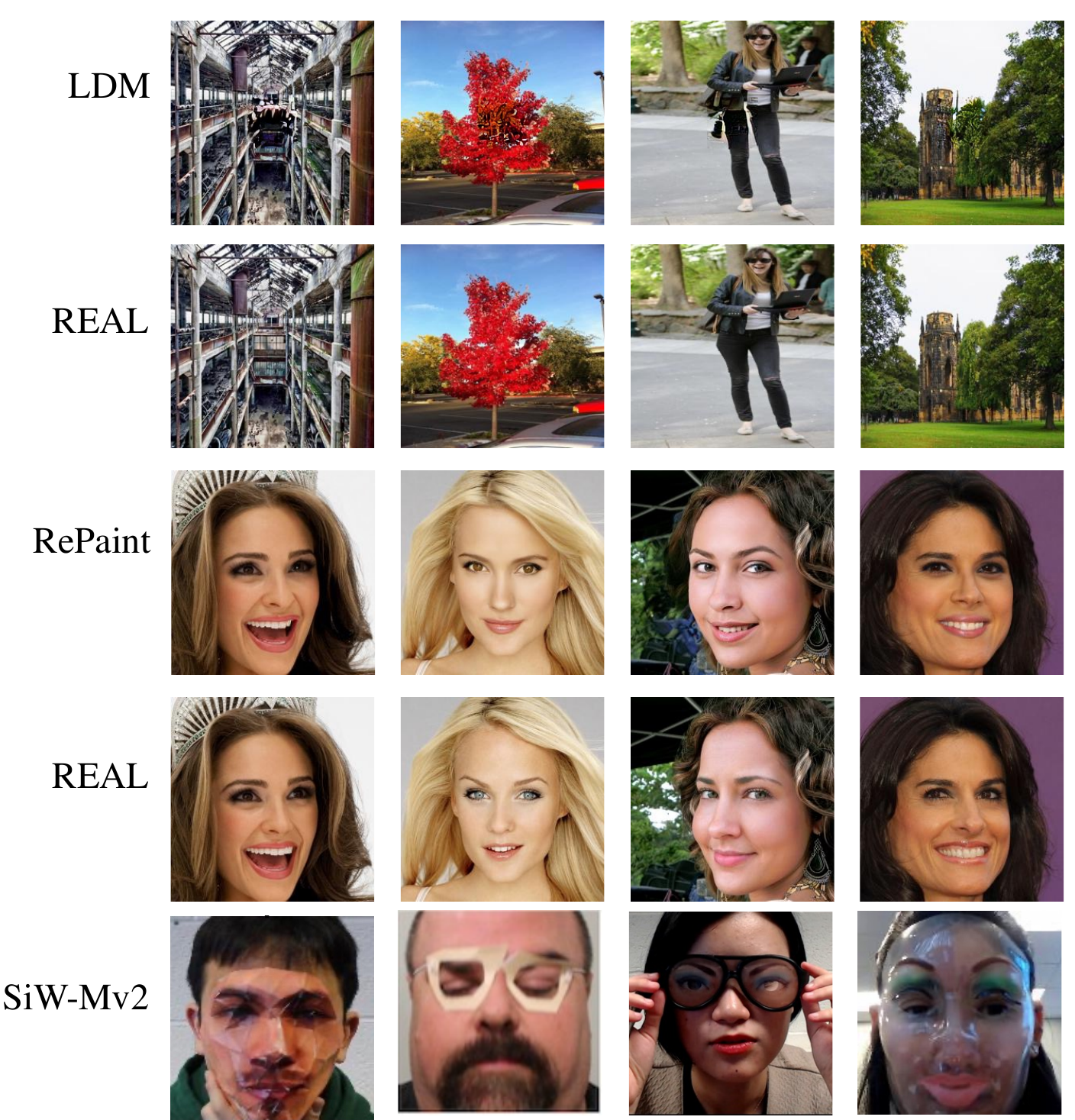}
    \caption{Images manipulated by LDM~\cite{rombach2021highresolution_latent_diffusion} and RePaint~\cite{lugmayr2022repaint} and their corresponding real images. The last row is the SiW-Mv$2$ dataset, where real human faces have spoof traces, such as funnyeyes and masks.}
    \label{fig:my_label}
\end{figure}

% In Fig.~\ref{fig_unseen_gen}, we evaluate the pre-trained model generalization ability on % on images generated by 
% unseen GAN and diffusion models, as well as real images from SiWM-v$2$ dataset~\cite{guo2022multi}, where facial image has spoof traces, such as funny glasses and wigs. Also, we measure the forgery localization performance on images manipulated by diffusion-based inpainting methods~\cite{rombach2021highresolution_latent_diffusion,lugmayr2022repaint}. 
Fig.~\ref{fig_unseen_gen} reports our method's generalization performance. Specifically, for each generative method and unseen domain real images, we have collected $1000$ images and use these images to form an inference dataset. After that, we apply the pre-trained HiFi-Net on such an inference to compute the classification accuracy, given $0.5$ fixed-threshold. 

Our first conclusion is the same as the most recent work~\cite{ricker2022towards} that some generative methods such as DSGAN~\cite{wang2022diffusion} and PNDM~\cite{liu2022pseudo} can generate rather sophisticated images that fool the powerful forgery detector. 

Secondly, we hypothesize that powerful forgery detector can largely fail when being applied on real images in different domain. For example, real images from SiW-Mv$2$ dataset~\cite{guo2022multi}, where facial image has spoof traces, such as funny glasses and wigs. 

Lastly, and more importantly, we observe the well-trained model always generalizes poorly on the image that is partially manipulated by diffusion model. We think this is because of two reasons: (1) conventional image editing methods are distinct by nature to the most recently proposed inpainting methods based on diffusion model; (2) the forgery area edited by diffusion model can have variations, not a rigid copy-move or removal manner that is commonly used by the traditional editing methods.

We believe these three aspects are valuable for the future research, namely: (a) how to make the model generalize well on detecting forged images created by advanced methods, (b) how to maintain the precision when we have real images from the new domain, (c) diffusion model based inpanting method can raise an issue for the existing forgery localization methods, indicating that new algorithm is needed.

\subsection{Image Editing Experiment}
\begin{figure*}[t]
    \centering
    \begin{overpic}[width=1\textwidth]{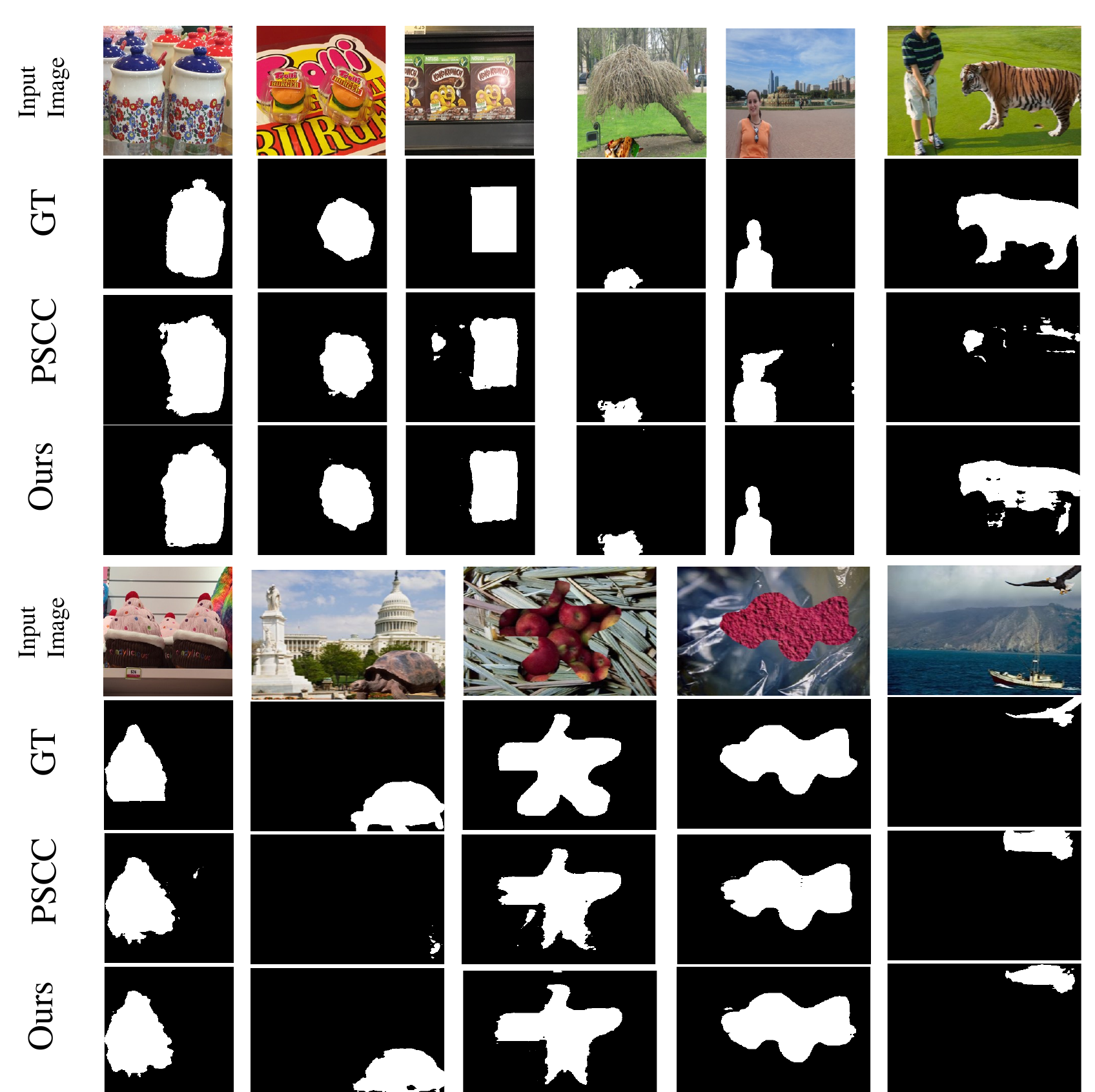}
    \end{overpic}
    \vspace{-2mm}
    \caption{Additional qualitative results on \textit{CASIA}, \textit{NIST\textit{16}} and \textit{Coverage} dataset.}
    \label{fig_supp_res_image_editing}
\end{figure*}

Following previous works~\cite{wang2022objectformer,liu2022pscc,hu2020span}, we evaluate the performance of our method against different post-processing steps, which is reported in Tab.~\ref{tab_editing_supp}. Our proposed method is more robust than the previous work, except for the post-processing of resizing $0.78$ times the image and JPEG compression with $50\%$ quality. Meanwhile, more qualitative results can be found in Fig.~\ref{fig_supp_res_image_editing}.
\begin{table}[ht]
    \footnotesize
    \centering
        \scalebox{1}{
        \begin{tabular}{c|cccc}
        \hline
        Post. &SPAN~\cite{hu2020span} &PSCC~\cite{liu2022pscc} &Obj.Fo.~\cite{wang2022objectformer} & Ours\\ \hline
        
        Resize ($0.78$)&$83.24$&$85.29$&$\mathbf{87.2}$&$86.9$\\
        Resize ($0.25$)&$80.32$&$85.01$&$86.3$&$\mathbf{86.5}$\\ \hline
        
        Gau.Blur ($3$)&$83.10$&$85.38$&$85.97$&$\mathbf{86.1}$\\ 
        Gau.Blur ($15$)&$79.15$&$79.93$&$80.26$&$\mathbf{81.0}$\\ \hline
        
        Gau.Noi ($3$)&$75.17$&$78.42$&$79.58$&$\mathbf{81.9}$\\ 
        Gau.Noi ($15$)&$67.28$&$76.65$&$78.15$&$\mathbf{79.5}$\\ \hline
        
        JPEG Co. ($100$)&$83.59$&$85.40$&$86.37$&$\mathbf{86.5}$\\ 
        JPEG Co. ($50$)&$80.68$&$85.37$&$\mathbf{86.24}$&$86.0$\\ \hline
        \end{tabular}
        }
    \vspace{-2mm}
    \caption{IFDL performance on \textit{NIST$\textit{16}$} with different post-processing steps. [Key: \textbf{Best}; Gau.: Gaussian; JPEG Co.: JPEG Compression.]\vspace{-3mm}.}
    \vspace{-1mm}
    \label{tab_editing_supp}
\end{table}

\subsection{The DFFD Dataset Performance}
We have included the complete version of our method performance on the DFFD dataset in Tab.~\ref{dffd_supp}. As we can see, compared to Attention Xception~\cite{stehouwer2019detection}, our method still achieves more accurate localization performance on Partial Manipulated and Fully Synthesized images. For the localization performance on the real images, our performance is comparable with the Attention Xception~\cite{stehouwer2019detection}.
\begin{table}[h]
    \centering
    %\footnotesize
    \begin{subtable}{.95\linewidth}
        \centering
        \resizebox{1\textwidth}{!}{
        \begin{tabular}{c|c c c}
            \hline
            IoU ($\uparrow$) / PBCA ($\uparrow$)& Real & Fu. Syn. & Par. Man.\\
            \hline
            Att.~\cite{stehouwer2019detection} &$-/\mathbf{0.998}$&$0.847/0.847$&$0.401/0.786$\\\hline
            Ours 
            &$-/0.978$&$\mathbf{0.893}/\mathbf{0.893}$&$\mathbf{0.411}/\mathbf{0.801}$ \\  \hline
        \end{tabular}}
        \caption{}
    \end{subtable}

    %\footnotesize
        \centering
    \begin{subtable}{.95\linewidth}
        \centering
        \resizebox{1\textwidth}{!}{
        \begin{tabular}{c|c c c}
            \hline
            IINC ($\downarrow$) / C.S. ($\downarrow$)& Real & Fu. Syn. & Par. Man.\\
            \hline
            Att.~\cite{stehouwer2019detection} &$0.015/-$&$0.077/\mathbf{0.095}$&$\mathbf{0.311}/0.429$\\\hline
            Ours 
            &$\mathbf{0.010}/-$&$\mathbf{0.060}/0.107$&$0.323/\mathbf{0.410
            }$ \\  \hline
        \end{tabular}}
    \caption{}
    \end{subtable}
    \vspace{-4mm}
    \caption{The localization performance: (a) Metrics are IoU and PBCA, the higher the better, (b) Metrics are IINC and Cosine Similarity, the lower the better. [Keys: Fu. Syn.: Fully-synthesized; Par. Man.: Partially-manipulated]}
    \label{dffd_supp}
\end{table}
\begin{figure*}[t]
    \centering
    \begin{overpic}[width=1.\textwidth]{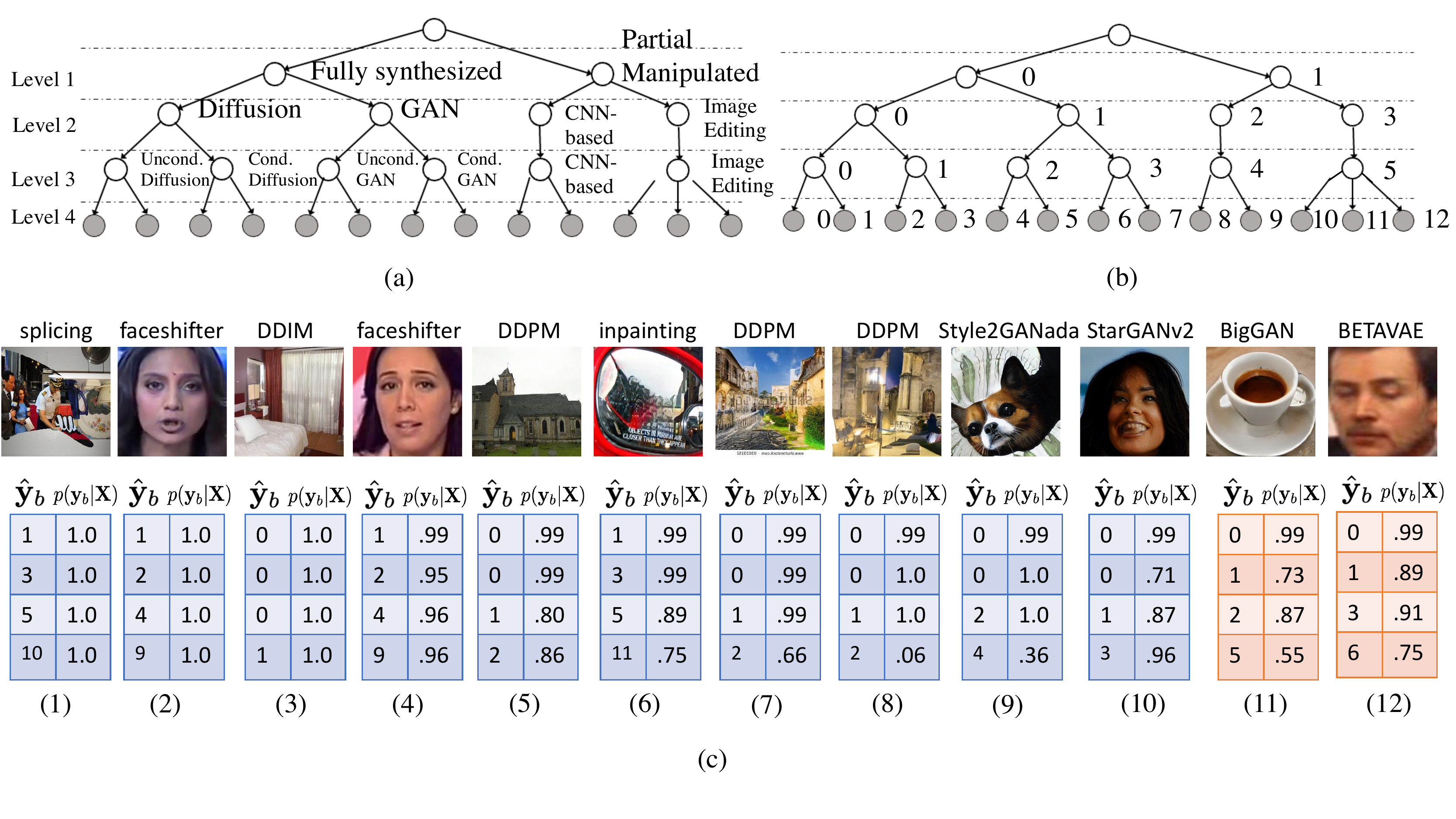}
    \end{overpic}
    \vspace{-10mm}
    \caption{(a) The original hierarchical structure used in the fine-grained classification at different levels. (b). For convenience, we assign category number to each forgery attribute, at different levels. (c). The detailed prediction probability for each input forged images. The table below each image reports the categorical value $\hat{\mbf{y}}_b$ and corresponding prediction probability $p(\mbf{y}_b|\mbf{X})$. From the top to down order, the results shown are for level$1$ to level$4$. For example, in the splicing image (example ($1$)): \texttt{partial manipulated$\rightarrow$image editing$\rightarrow$image editing$\rightarrow$splicing}, which corresponds to the label index $1,3,5,10$ that are shown in the first table. Lastly, the example ($1$) -- ($10$) are seen forgery method in the training, and the example ($11$) and ($12$) are unseen forgery attribute.}
    \label{fig_probability}
\end{figure*}

\subsection{Forgery Attribute Classification}
We have included specific classification results for a variety of samples. 
Tab.~(6) of the paper reports the fine-grained classification result of HiFi-IFDL. Here we show the classification probability at different levels. In examples ($5$) and ($10$) of Fig.~\ref{fig_probability}, we can see the robustness of our proposed method that learns the hierarchical structure. The $3$rd level fine-grained prediction probability on $5$th example and $10$th example is lower than the fine-grained classification prediction probability on the $4$th level. This means our algorithm can recover the accuracy at the fine level classification even the classification on the coarser level does not perform excellent.

\subsection{Implementation Details}
% We have constructed our HiFi-Net based on the pre-trained HR-Net.
% Specifically, as depicted in Fig.~\ref{fig_implementation}, we take the branch $\net_{1}$ directly from the HR-Net, and $\net_{2}$, $\net_{3}$ and $\net_{4}$ are duplicating the same convolution layers from the HR-Net. 
% in which we make the convolution layer number on each branch same.
In our HiFi-Net, the feature map resolution for different branches are $256$, $128$, $64$, and $32$ pixels. In the experiment on HiFi-IFDL, the fine-grained classification for $1$st, $2$nd, $3$rd and $4$th levels are $2$-way, $4$-way, $6$-way and $14$-way multi-class classification, respectively. The $3$rd level fine-grained classification categories are: \texttt{unconditional diffusion}, \texttt{conditional diffusion}, \texttt{unconditional GAN}, \texttt{conditional GAN}, \texttt{CNN-based partial manipulation} and \texttt{Image editing}. 

As for the details of $\mathcal{L}_{loc}$ implementation, we first use the initialized HiFi-Net to convert each pixel in the input image to the high-dimensional feature $\mathbf{F}^{\prime}_{ij} \in R^{D}$, where $D = 18$. Then we average the feature $\mathbf{F}^{\prime}_{ij} \in R^{D}$ for all pixels in the real image from the HiFi-IFDL, and this average value then is used as $\mathbf{c}$. Then, we compute the $\ell_{2}$ distance between each pixel feature $\mathbf{F}^{\prime}_{ij} \in R^{D}$ and $\mathbf{c}$, and denote the largest distance as $D_{max}$. In the Eq.~(1) of the paper, we set the threshold $\tau$ as $2.5\cdot D_{max}$.

The architecture is trained end-to-end with different learning rates per layers. The detailed objective function is:
\begin{equation*}
\small
\mathcal{L}_{tot} = 
    \begin{cases} 
        $100*$\mathcal{L}_{loc} + \mathcal{L}^{1}_{cls} + \mathcal{L}^{2}_{cls} + \mathcal{L}^{3}_{cls} + $100*$\mathcal{L}^{4}_{cls} & \mbox{if $\mbf{X}$ is forged}\\ 
        
        \mathcal{L}_{loc} + \mathcal{L}^{4}_{cls} & \mbox{if $\mbf{X}$ is real}
    \end{cases}
\end{equation*}
Where $\mbf{X}$ is the input image. When the input image is labeled as ``real'', we only apply the last branch ($\net_{4}$) loss function, otherwise, if it is labeled as ``manipulated'', we use all the branches. 

The entire architecture is trained for $13$ epoches, and all training samples are seen by the model in each epoch. The feature extractor is modified based on the pre-trained HRNet~\cite{wang2020deepHRNet}, in which we add more layers such that each branch of our multi-branch feature extractor can have identical number of convolutional layers, and the details can be found in our source code that will be released upon the acceptance. We use $\texttt{1e-4}$ to train the multi-branch feature extractor and classification module, and $\texttt{3e-4}$ to train the localization modules. During the training, we use $\texttt{ReduceLROnPlateau}$ as the learning rate scheduler to reduce the learning rate.
% In the multi-branches feature extractor, the convolution layers, which are highlighted in Fig.~\ref{fig_implementation}, are trained with a learning rate of $\texttt{4e-5}$ and the rest of the multi-branch feature extractor with  $\texttt{5e-5}$. 

\subsection{Societal Impact}
Our work has the positive societal impact to the community. Because our work is dealing with various categories of forgery methods, which enable the algorithm to detect all kinds of manipulation, including seen and unseen forgeries, as indicated by Supplementary section 4. Our algorithm can enable a tool that makes general public in our society to have more trust in media contents.

\clearpage